\documentclass[runningheads]{llncs}
\usepackage{graphicx}

\usepackage{tikz}
\usepackage{comment}
\usepackage{amsmath,amssymb} 
\usepackage{color}
\usepackage{caption}
\usepackage{subcaption}
\usepackage{multirow}
\usepackage{makecell}
\usepackage{threeparttable}
\usepackage{booktabs}
\usepackage{float}
\usepackage{hyperref}
\usepackage{wrapfig}
\usepackage{enumerate}

\hypersetup{	
  hidelinks,	
  colorlinks=true,
  allcolors=magenta,
}

\definecolor{mygray}{gray}{.9}

\usepackage{color, colortbl}

\newcommand{\mycomment}[1]{}
\newcommand{\TODO}[1]{\textcolor{teal}{[#1]}}

\newcommand{\zc}[1]{\textcolor{brown}{[#1]}}
\newcommand{\xz}[1]{\textcolor{red}{[#1]}}
\newcommand{\hc}[1]{\textcolor{orange}{[#1]}}

\newcommand\redout{\bgroup\markoverwith{\textcolor{red}{\rule[0.5ex]{1pt}{1.5pt}}}\ULon}

\newcommand{\removed}[1]{\textcolor{red}{\sout{#1}}}

\newif\ifclean			  

\usepackage{amssymb}
\usepackage{pifont}%
\ifclean
	\renewcommand{\TODO}[1]{}	
	\renewcommand{\hc}[1]{}
	\renewcommand{\xz}[1]{}
	\renewcommand{\zc}[1]{}
	
	\renewcommand{\st}[1]{}
	\renewcommand{\redout}[1]{}

	\renewcommand{\removed}[1]{}
	
\fi

\newcommand\eqscale{0.9}

\begin{document}
\pagestyle{headings}
\mainmatter
\def\ECCVSubNumber{1666}  

\title{Physical Attack on Monocular Depth Estimation with Optimal Adversarial Patches} 

%
\author{Zhiyuan Cheng \inst{1}\and
James Liang \inst{2} \and
Hongjun Choi\inst{1} \and
Guanhong Tao\inst{1} \and
Zhiwen Cao\inst{1} \and
Dongfang Liu\inst{2} \and
Xiangyu Zhang\inst{1}}
\authorrunning{Z. Cheng, J. Liang et al.}
%
\institute{Purdue University \and
Rochester Institute of Technology \\ 
\email{\{cheng443, choi293, taog, cao270, xyzhang\}@cs.purdue.edu}\\
\email{\{jcl3689, dongfang.liu\}@rit.edu}
}
\maketitle
\vspace{-10pt}
\begin{abstract}
Deep learning has substantially boosted the performance of Monocular Depth Estimation (MDE), a critical component in fully vision-based autonomous driving (AD) systems ($e.g.$, Tesla and Toyota). In this work, we develop an attack against learning-based MDE. 
In particular, we use an optimization-based method to systematically generate stealthy physical-object-oriented adversarial patches to attack depth estimation. We balance the stealth and effectiveness of our attack with object-oriented adversarial design, sensitive region localization, and natural style camouflage.
Using real-world driving scenarios, we evaluate our attack on concurrent MDE models and a representative downstream task for AD ($i.e$., 3D object detection). Experimental results show that our method can generate stealthy, effective, and robust adversarial patches for different target objects and models and achieves more than 6 meters mean depth estimation error and 93\% attack success rate (ASR) in object detection with a patch of 1/9 of the vehicle's rear area.
Field tests on three different driving routes with a real vehicle indicate that we cause over 6 meters mean depth estimation error and reduce the object detection rate from 90.70\% to 5.16\% in continuous video frames.

\vspace{-5pt}
\keywords{Physical Adversarial Attack, Monocular Depth Estimation, Autonomous Driving.}
\end{abstract}

\vspace{-30pt}
\section{Introduction}
\vspace{-5pt}

Monocular$_{\!}$ Depth$_{\!}$ Estimation$_{\!}$ (MDE) is$_{\!}$ a technique$_{\!}$ for$_{\!}$ estimating$_{\!}$ the$_{\!}$ distance$_{\!}$ between an object and the camera$_{\!}$ from$_{\!}$ RGB$_{\!}$ image$_{\!}$ inputs.$_{\!}$ It$_{\!}$ is$_{\!}$ a critical$_{\!}$ vision$_{\!}$ task$_{\!}$ for$_{\!}$ autonomous driving$_{\!}$ (AD)$_{\!}$ because it bridges the gap between$_{\!}$ Lidar$_{\!}$ sensors$_{\!}$ and$_{\!}$ RGB$_{\!}$ cameras~\cite{wang2019pseudo}$_{\!}$ and$_{\!}$ its$_{\!}$ measurement$_{\!}$ has$_{\!}$ an$_{\!}$ effect$_{\!}$ on $_{\!}$a variety of downstream perception tasks ($e.g.,$ object$_{\!}$ detection~\cite{wang2019pseudo},$_{\!}$ visual SLAM~\cite{wimbauer2020monorec},$_{\!}$ and visual relocalization~\cite{lm-reloc-2020}). For$_{\!}$ its$_{\!}$ importance,$_{\!}$ Tesla$_{\!}$ has$_{\!}$ integrated$_{\!}$ MDE$_{\!}$ into$_{\!}$ its$_{\!}$ production-grade$_{\!}$ Autopilot$_{\!}$ system~\cite{tesla-AI-Day,tesla-self-supervised},$_{\!}$ and$_{\!}$ other$_{\!}$ AD$_{\!}$ companies$_{\!}$ such$_{\!}$ as$_{\!}$ Toyota~\cite{packnet} and Huawei~\cite{aich2020bidirectional} are also actively investigating this technique. 
With the increasing popularity of MDE, ensuring its security becomes a prominent challenge.

Existing 
adversarial$_{\!}$ attacks$_{\!}$ against MDE$_{\!}$ are implemented in digital-~\cite{zhang2020adversarial,wong2020targeted}$_{\!}$ or$_{\!}$ physical-world platforms~\cite{yamanaka2020adversarial}.$_{\!}$ Compared$_{\!}$ to$_{\!}$ digital-world attacks,$_{\!}$ attacks$_{\!}$ in$_{\!}$ the$_{\!}$ physical$_{\!}$ world$_{\!}$ are$_{\!}$ more$_{\!}$ challenging$_{\!}$ because$_{\!}$ they$_{\!}$  require$_{\!}$ robust$_{\!}$ perturbations$_{\!}$ to overcome$_{\!}$ various$_{\!}$ photometric$_{\!}$ and$_{\!}$ geometric$_{\!}$ changes~\cite{athalye2018synthesizing},$_{\!}$ reducing$_{\!}$ their$_{\!}$ stealth.$_{\!}$ Prior$_{\!}$ efforts$_{\!}$ for$_{\!}$ physical-world$_{\!}$ adversarial attacks~\cite{yamanaka2020adversarial,song2018physical,komkov2021advhat,brown2017adversarial}$_{\!}$ generally$_{\!}$ employ$_{\!}$ an unnatural-looking$_{\!}$ adversarial$_{\!}$ patch$_{\!}$ and$_{\!}$ sacrifice$_{\!}$ stealth$_{\!}$ for$_{\!}$ attack$_{\!}$ effectiveness,$_{\!}$ leaving plenty of room for improvement. Additionally, with MDE's rapid development, many$_{\!}$ downstream$_{\!}$ tasks$_{\!}$ that$_{\!}$ previously$_{\!}$ require$_{\!}$ expensive$_{\!}$ Lidar$_{\!}$ sensors$_{\!}$ or depth$_{\!}$ cameras$_{\!}$ can now$_{\!}$ be performed entirely with MDE's measurement$_{\!}$ and achieve$_{\!}$ competitive$_{\!}$ performance. However, the investigation of the impact of compromised MDE on these downstream tasks remains largely unknown.

To address the aforementioned problems, in this paper, we investigate the \textit{stealth of physical-world attack against MDE} and present a physical-object-oriented adversarial patch optimization framework to generate \textit{stealthy, effective and robust  adversarial patches} for target objects ($e.g.$, vehicles and pedestrians). In particular, we are able to achieve the followings: \ding{182} we design a physical-object-oriented adversarial optimization, which binds the patch and the target object together regarding attack effects and physical-world transformations (\S\ref{sec:adv_gen});
\ding{183} we optimize the patch region on the target object with a differentiable patch mask representation, which automatically locates the highly effective area for attack on the target object and improves attack performance with a small patch size (\S\ref{sec:sen_loc}); \ding{184} we camouflage the adversarial pattern with natural styles ($e.g.$, rusty and dirty) with deep photo style transfer~\cite{luan2017deep}, resulting in stealthier patch for the attack (\S\ref{sec:atk_cam});  \ding{185} we investigate the impact of compromised MDE on a representative downstream task in AD --- 3D object detection (\S\ref{sec:downstream}).  Fig.~\ref{fig:highlight} gives an example. In addition, we examine our attack with popular defence techniques (\S\ref{sec:defence}). Our key contributions are:

\begin{figure}[t]
    \centering
    \includegraphics[width=\textwidth]{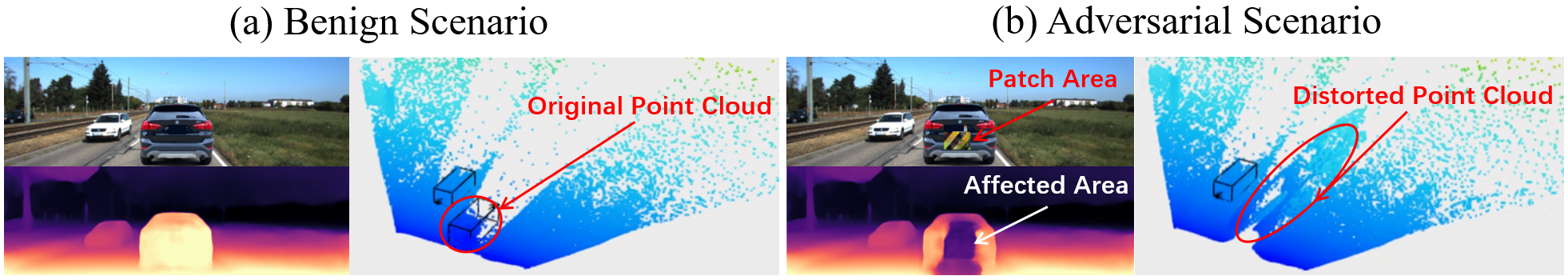}
    \vspace{-20pt}
    \caption{\small Attack MDE and 3D object detection with a natural adversarial patch. The left is a benign scenario and the right is the corresponding adversarial scenario. 3D object detection takes the pseudo-Lidar ($i.e.$, point cloud projected from 2D depth map) as input and outputs bounding boxes of recognized objects. Observe in the adversarial scenario (b) that our optimized adversarial patch can disturb the depth estimation 
    of the target vehicle significantly and the effect propagates to an area larger than the patch itself. 
    Pseudo-Lidar of the vehicle is thus distorted and it cannot be detected in the downstream task. }
    \label{fig:highlight}
    \vspace{-15pt}
\end{figure}

\vspace{-5pt}
\begin{enumerate}
    \item We develop a physical-object-oriented adversarial patch attack against MDE that balances stealth and effectiveness. To the best of our knowledge, we are the first to investigate stealthy physical-world attacks against MDE considering both the patch size and naturalness.
    \item  We propose an optimization framework that considers physical object characteristics, 
     has a differentiable patch region representation, and provides natural style based camouflage.
    \item  We evaluate our attack on 3 representative MDE models and a downstream task with real-world driving scenarios in both digital and physical worlds. Our attack is effective on different types of target objects and state-of-the-art models. It causes over 6 meters of mean depth estimation error for a real vehicle, with a patch only 1/9 of the vehicle's rear area, and achieves more than 90\% attack success rate in 3D object detection. A video is available at \url{https://youtu.be/L-SyoAsAM0Y}.
\end{enumerate}

\vspace{-15pt}
\section{Related Work}
\vspace{-5pt}
\textbf{AD Systems Security.} In AD, sensor security and autonomy software security are the two important challenges. For sensor security, prior works focus on spoofing/jamming on camera~\cite{yan2016can,nassi2020phantom,petit2015remote}, LiDAR~\cite{cao2019adversarial,shin2017illusion}, RADAR~\cite{yan2016can}, ultrasonic~\cite{yan2016can}, GPS~\cite{shen2020drift} and IMU~\cite{tu2018injected,trippel2017walnut}. For autonomy software security, some prior works study regression tasks ($e.g.,$ depth estimation~\cite{yamanaka2020adversarial} and optical flow estimation~\cite{ranjan2019attacking}), and others focus on classification tasks ($e.g.,$ 2D object detection and classification~\cite{song2018physical,brown2017adversarial}, tracking~\cite{jia2020fooling}, lane
detection~\cite{sato2020hold,sato2021wip}, and traffic light detection~\cite{tang2021fooling}). This work focuses on autonomy software security, that is, compromising MDE and its related downstream tasks. 

\noindent
\textbf{Physical-world Adversarial Attacks.} Many prior efforts in adversarial attacks have been directed toward generating patches or perturbations in the digital space~\cite{pei2017deepxplore,goodfellow2014explaining,papernot2016limitations,xiao2018generating,xiao2018spatially,xiao2019characterizing,qiu2020semanticadv,tsai2020robust,xiao2019meshadv}. In comparison, we conduct extensive experiments on adversarial attacks in the physical world. Although existing physical-world attacks have addressed tasks such as image classification~\cite{song2018physical,brown2017adversarial}, object detection~\cite{chen2018shapeshifter,xu2020adversarial,thys2019fooling}, face recognition~\cite{sharif2016accessorize,komkov2021advhat}, the domain of depth estimation attack has received scant attention. Moreover, the correlations between stealth and attack effectiveness are largely understudied in the literature. In this paper, we make an attempt to close the aforementioned knowledge gap.

\noindent
\textbf{MDE Attacks.} 
Zhang~\cite{zhang2020adversarial} proposes a multi-task attack strategy to improve the performance in the universal attack scenario. Wong~\cite{wong2020targeted} proposes a way to generate targeted adversarial perturbation on images and alter the depth map arbitrarily. These two attacks focus on digital-space perturbations thus are not directly applicable in the physical world. Yamanaka~\cite{yamanaka2020adversarial} proposes a method to generate printable adversarial patch for MDE but it does not consider stealth of the patch. Different from prior efforts, we focus on the stealth and to the best of our knowledge, we are the first to examine the stealth of adversarial patches for physical-world attack against MDE.

\vspace{-10pt}
\section{Method}\label{sec:method}
\vspace{-5pt}

\begin{figure}[t]
    \centering
    \includegraphics[width=\textwidth]{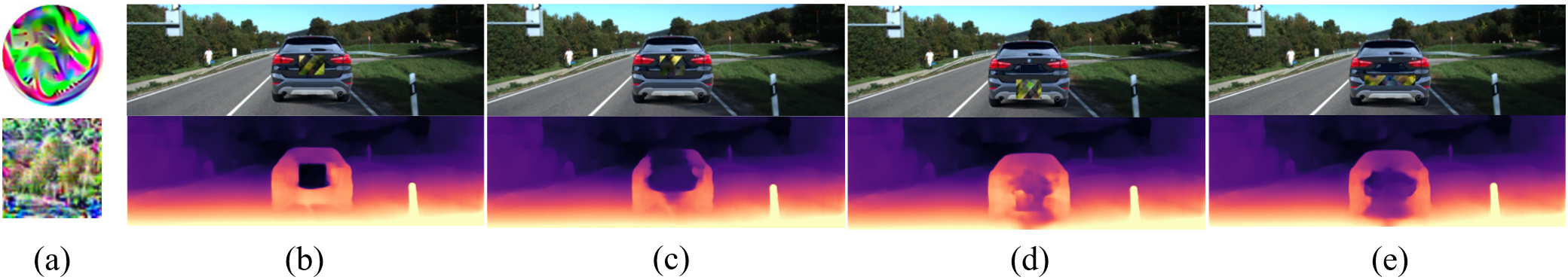}
    \vspace{-20pt}
    \caption{\small (a): Unconstrained adversarial patches in~\cite{yamanaka2020adversarial,mathew2020monocular} are easy to be identified; traditional patch-oriented attack in (b) affects smaller area than our object-oriented attack in (c); (c), (d) and (e): different regions on the target object have different sensitivity regarding attack effect even with the same total area.\vspace{-18pt}}
    \label{fig:motivation}
\end{figure}

\subsection{Physical-object-oriented MDE Attack}
\vspace{-5pt}
\subsubsection{Motivation}
Compared with unconstrained adversarial patches (see Fig.~\ref{fig:motivation}a) which often look suspicious, stealthy patches may draw less attention and hence can stay on  the target vehicle for an extended period of time, posing a greater threat. We divide the challenge of achieving stealth into two sub-problems: patch size minimization and achieving natural appearance. 
To minimize patch size, we investigate how to maximize the attack effect with smaller patches and propose two approaches: \ding{182} enlarging the patch's affected area (see comparison in Fig.~\ref{fig:motivation}b and c), and \ding{183} locating the adversarial patch in a more sensitive region of the target vehicle (see Fig.~\ref{fig:motivation}c, d and e). In terms of naturalness, as the magnitude of perturbations required to launch attack in the physical world is much more substantial, we cannot simply bound the adversarial noise to a human un-noticeable level via various $L_p$-norms as in digital-world attacks, which provides little physical-world robustness.
Instead, we use style transfer to disguise the adversarial pattern as natural styles ($e.g.,$ dirty or rusty).
\vspace{-15pt}
\subsubsection{Attack Pipeline}
\begin{figure}[t]
    \centering
    \includegraphics[width=\textwidth]{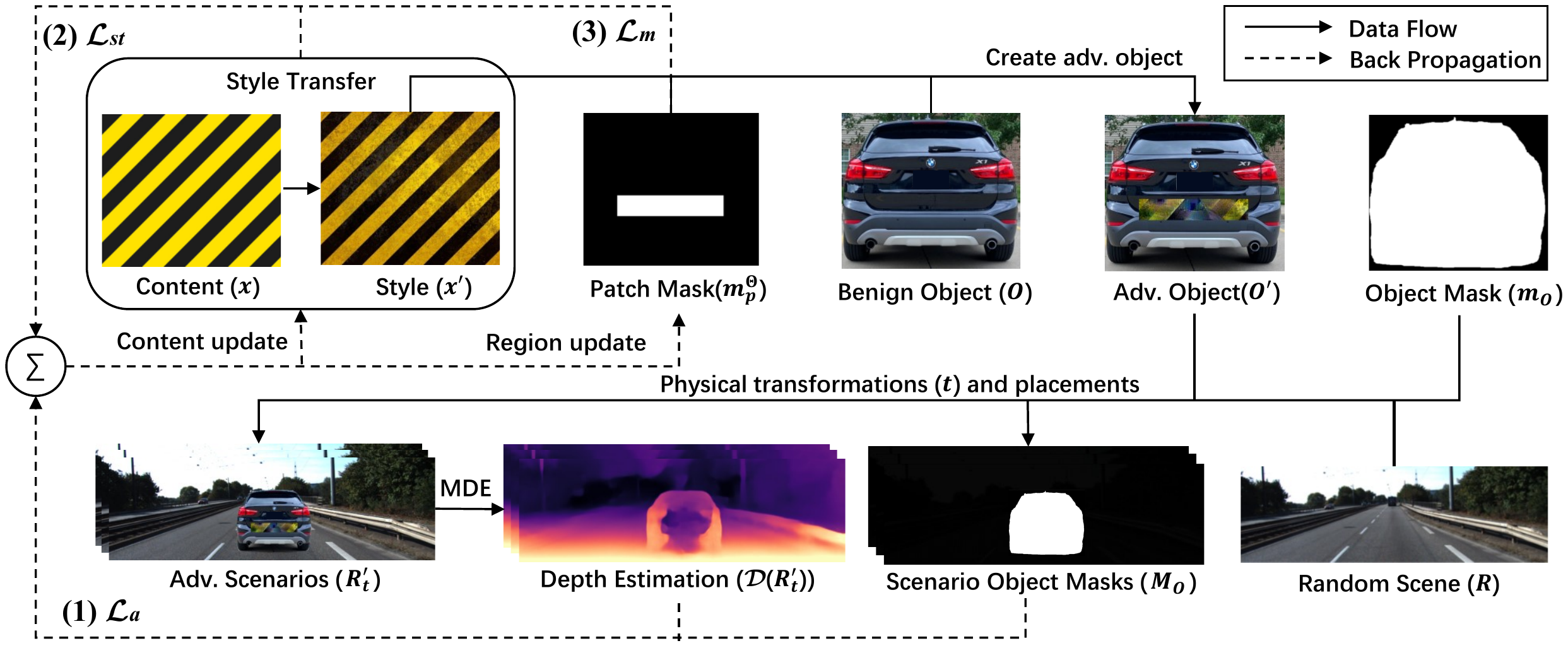}
    \caption{Overview of the physical-object-oriented framework to generate a stealthy adversarial patch. }
    \label{fig:overview}
    \vspace{-10pt}
\end{figure}

We use an optimization-based method to generate adversarial patches and there are three main optimization goals:  \ding{182} increasing the estimated distance of target object (\S\ref{sec:adv_gen});  \ding{183} minimizing the patch to locate a sensitive ($i.e.,$ most effective) region for attack (\S\ref{sec:sen_loc}), and  \ding{184} camouflaging the adversarial patch with natural styles (\S\ref{sec:atk_cam}). Fig.~\ref{fig:overview} shows the overview of our attack. From the top left, we start with style transfer on the patch content image. Next, we crop the style-transferred patch with an optimizable patch mask ($m_p^\Theta$) and paste it onto a target object ($O$) ($e.g.,$ a vehicle) creating an adversarial one ($O'$). Then, we synthesize adversarial scenarios ($R'_t$) by placing the adversarial object into random scenes with physical transformations ($t$) and estimate scenarios' depth ($\mathcal{D}(R'_t)$). We define an adversarial loss ($\mathcal{L}_a$) to increase depth of the target object. Together with a style transfer loss ($\mathcal{L}_{st}$) maintaining the naturalness and a patch size loss ($\mathcal{L}_m$) minimizing the patch, we perform back propagation and update the patch content and the mask iteratively to address the three optimization goals. The solid lines denote data flow and the dashed lines represent back propagation paths. Each component is  explained in details in the following sections.

\vspace{-10pt}
\subsection{Adversarial Perturbation Generation}\label{sec:adv_gen}
\vspace{-5pt}

In preparation, we take a photo of the target object ($O$) and select a patch content image ($x$) and a style image. Given the patch mask ($m_p$), we create an adversarial object ($O'$) by applying the style-transferred patch ($x'$) on the benign object in the following way:
\begin{gather}
\scalebox{\eqscale}{$
\begin{aligned}
    O' = O \odot (1-m_p) + x' \odot m_p,
\end{aligned}
$}
\end{gather}
where $\odot$ denotes the element-wise multiplication and $O, m_p, x'$ have the same width and height. We explain the patch mask definition and style transfer later in \S\ref{sec:sen_loc} and \S\ref{sec:atk_cam}. We evaluate the depth of the target object inside a scene because the camera on the victim vehicle captures scene frames as input instead of independent objects. Specifically, in each optimization iteration, we randomly sample a scene from the dataset and paste the adversarial object into the scene to create an adversarial scenario. Unlike previous attacks against autonomous driving systems~\cite{cao2021invisible,sato2021dirty} that aim at a particular scene or a road section, our attack is universal and scene-independent. 

\begin{figure}[t]
	\centering
	\begin{minipage}{.37\columnwidth}
		\centering
        \includegraphics[width=1.0\columnwidth]{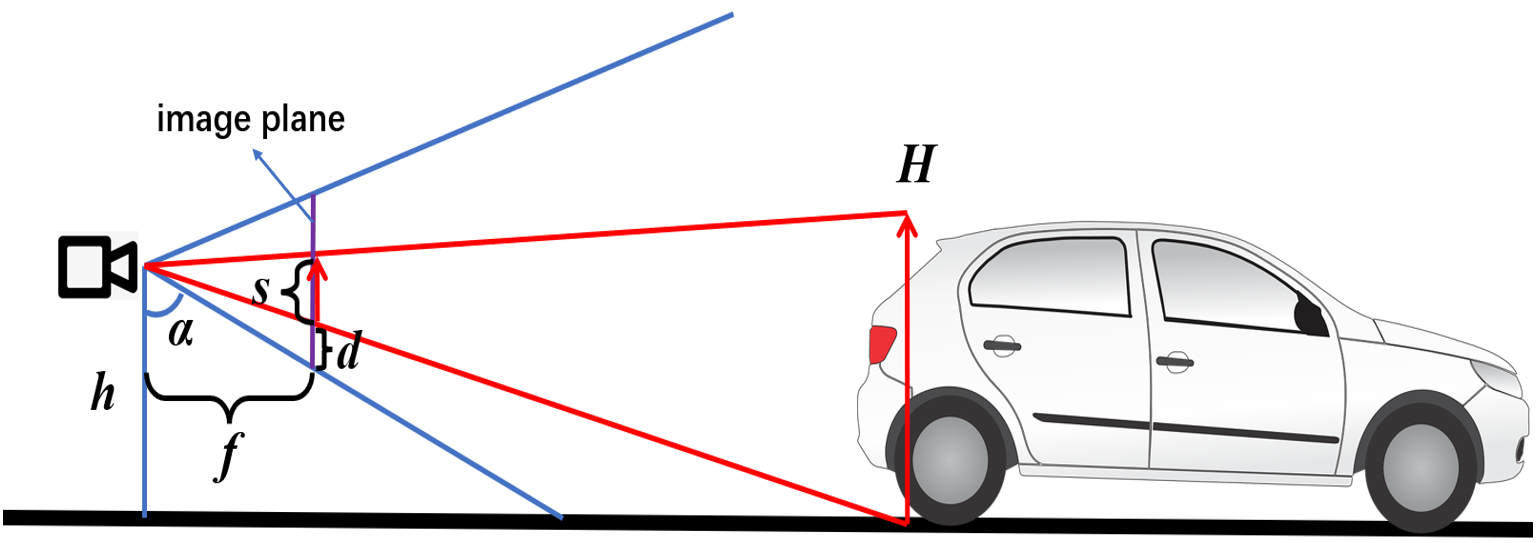}
        \vspace{0pt}
        \caption{\small Perspective projection of a vehicle (side view).}
        \label{fig:Camera}
	\end{minipage}%
	\begin{minipage}{.60\columnwidth}
		\centering
        \includegraphics[width=1\columnwidth]{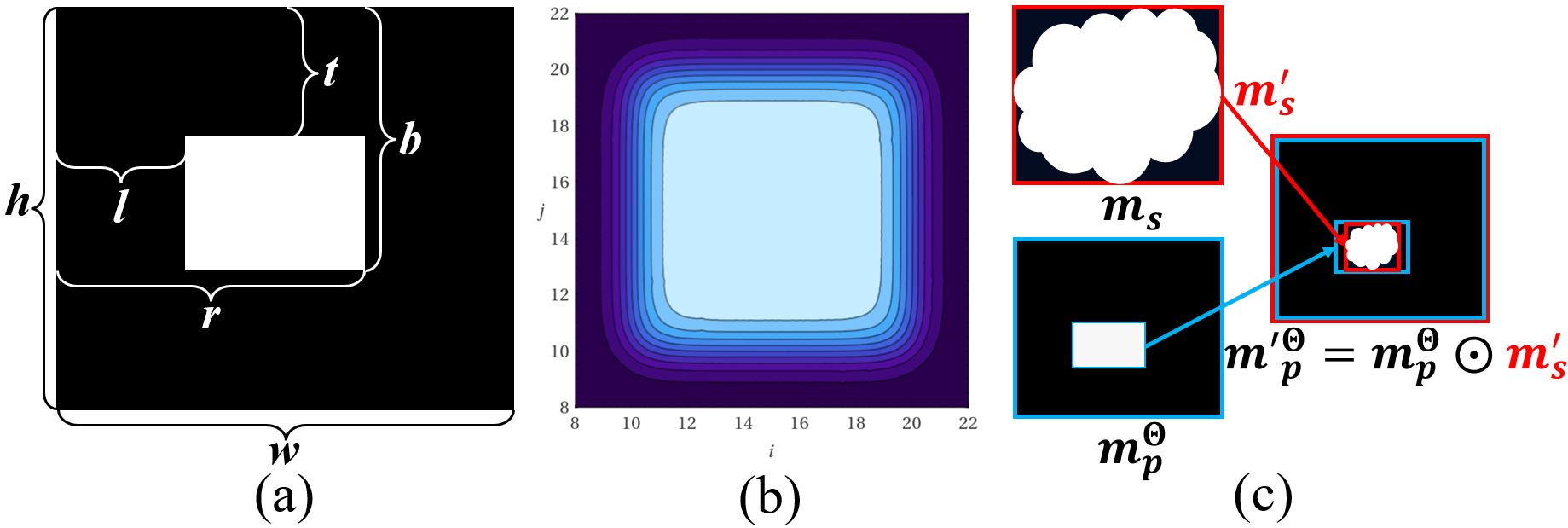}
        \vspace{-15pt}
        \caption{ \small Patch region definitions.}
        \label{fig:mask_def}
	\end{minipage}%
	\vspace{-10pt}
\end{figure}

To improve the robustness of our attack in the physical world, we apply Expectation of Transformation (EoT) \cite{athalye2018synthesizing} by randomly transforming the object in size, rotation, brightness, saturation, etc., before pasting. The horizontal position of pasting is random, while the vertical position is calculated according to the size of the object considering physical constraints. Specifically, Fig.~\ref{fig:Camera} shows the perspective model of a vehicle in a side view and we assume the camera is facing straight forward without tilt. $H$ is the height of the target vehicle; $h$ is the height of the camera with respect to the victim vehicle; $f$ is the focus length of the camera and $\alpha$ relates to the camera's view angle. On the image, the vertical position of the vehicle ($d$) is calculated from the height of the vehicle ($s$) with Equation~\ref{eq:perspective}. Intuitively, objects farther away appear smaller in perspective so a smaller object after transformation is pasted to a higher vertical position on the image, which is closer to {\em the vanishing point} (of the camera), which denotes the furthest physical point in the camera view, 
and has further depth estimation. 
\vspace{-10pt}
\begin{gather}\label{eq:perspective}
\scalebox{\eqscale}{$
\begin{aligned}
    d = -\frac{h}{H}s + \frac{f}{\tan\alpha}
\end{aligned}
$}
\end{gather}

Formally, the adversarial scenario 
$R'_t$ is described with Equation~\ref{eq:adv_scene},
\begin{gather}\label{eq:adv_scene}
\scalebox{\eqscale}{$
\begin{aligned}
    R'_t = \Lambda_t\left(t(O' \odot m_o), R\right)
\end{aligned}
$}
\end{gather}
where $t$ is the random transformation applied on the target object; $m_o$ is the object mask used to extract the object from the image; $R$ is the randomly sampled scene from database and $\Lambda(\cdot, \cdot)$ is the paste operation to combine an adversarial object and a scene following the physical constraint in Equation~\ref{eq:perspective}. Since our goal is to make the target object further away, we want to maximize the object's depth estimation ($i.e.,$ minimize the reciprocal). Hence, we define the adversarial loss in Equation~\ref{eq:adv_loss}, where $T$ is a set of transformations; $D_R$ is a set of scenes; $MSE(\cdot, \cdot)$ is the mean square error between two variables; $\mathcal{D}$ is the depth estimation model and $M_o$ is the object mask in the scenario. 
\vspace{-5pt}
\begin{gather}\label{eq:adv_loss}
\scalebox{\eqscale}{$
\begin{aligned}
    \mathcal{L}_a =
    \mathbf{E}_{t\sim T, R\sim D_R} \left[MSE\left(\mathcal{D}\left(R'_t\right)^{-1} \odot M_o, 0\right) \right]
\end{aligned}
$}
\end{gather}

\vspace{-15pt}
\subsection{Sensitive Region Localization}\label{sec:sen_loc}

As described in \S\ref{sec:adv_gen}, we apply the style transferred patch $x'$ onto the target object by a patch mask $m_p$ which defines the patch region on the target object. 
Prior works~\cite{wang2019neural-NC,liu2019abs,lee2021bbam}  optimizing masks treat each pixel of the mask as a parameter and the generated mask suffers from low deployability due to sparse and scattered mask regions (See Fig.~\ref{fig:depth_baseline_mask}b). Instead we design a novel rectangular patch region optimization method (we call it \textit{regional optimization}) to locate a sensitive region automatically. Although we define the patch region as rectangular, the final patch is not necessarily rectangular but have an arbitrary predefined shape. Details are explained later.

A typical rectangular patch mask has ones within the rectangular borders and zeros otherwise.
However, this mask is not differentiable regarding the border parameters because the mask values are not continuous across the borders and border information is not encoded into each mask values, which means that the region cannot be optimized via gradient descent and back propagation. To solve this problem, we design a differentiable soft version of the rectangular mask making it optimizable with respect to four border parameters. 
Specifically, we define border parameters $\Theta = [l, r, t, b]$ as shown in Fig.~\ref{fig:mask_def}a. $l$ and $r$ are the left and right borders' column indices and $t$ and $b$ are the top and bottom borders' row indices. Let $w$ and $h$ be the width and height of the mask respectively and we have $0\leq l \leq r \leq w$ and $0 \leq t \leq b \leq h$. 
\vspace{-5pt}
\begin{gather}\label{eq:old_mask}
\scalebox{\eqscale}{$
\begin{aligned}
    m_p^{\Theta} = \{m_p^{\Theta}[i,j]~|~i \in {1...w}, j \in {1...h}\}\\
    m_p^{\Theta}[i,j] = \frac{1}{4}(-sign(i-t)\cdot sign(i-b) + 1)\\\cdot (-sign(j-l)\cdot sign(j-r) + 1),
\end{aligned}
$}
\end{gather}
Typically, a mask is defined by Equation~\ref{eq:old_mask} with $\Theta$ as parameters, where $m_p^\Theta  \in \{0, 1\}^{w\times h}$ is the patch mask and $[i,j]$ is index of the pixel at $i$-th row and $j$-th column; $sign(x)$ outputs $1$ when $x\geq0$ and $-1$ when $x<0$; and $m_p^{\Theta}[i,j]$ evaluates to one if and only if the pixel is within the four borders defined by $\Theta$ and zero otherwise. To make each mask value differentiable regarding border parameters and maintain the property of original definition, we approximate $sign(\cdot)$ by $\tanh(\cdot)$ and define the patch mask with Equation~\ref{eq:soft_mask}.
\vspace{-5pt}
\begin{gather}
\scalebox{\eqscale}{$
\begin{aligned}\label{eq:soft_mask}
    m_p^{\Theta}[i,j] = \frac{1}{4}(-\tanh(i-t)\cdot \tanh(i-b) + 1)\\\cdot (-\tanh(j-l)\cdot \tanh(j-r) + 1)
\end{aligned}
$}
\end{gather}
Fig.~\ref{fig:mask_def}b is an example of the mask defined by us. In this example, $w$ and $h$ are 30, $l$ and $t$ are 10, and $r$ and $b$ are 20. Observe that the borders of the rectangular region change gradually. Each pixel value is encoded with border parameters $\Theta$.

In the beginning, the patch mask is initialized to cover the whole image, ($i.e.,$$_{\!}$ $l_{\!} =_{\!} t_{\!} =_{\!} 0$, $b_{\!} =_{\!} h$ and $r_{\!} =_{\!} w$).$_{\!}$ One$_{\!}$ of$_{\!}$ our$_{\!}$ optimization$_{\!}$ goal$_{\!}$ is$_{\!}$ to$_{\!}$ minimize the mask area, thus we define a mask loss term (Equation~\ref{eq:mask_loss}) to penalize the area of mask. 
\vspace{-5pt}
\begin{gather}\label{eq:mask_loss}
\scalebox{\eqscale}{$
\begin{aligned}
    \mathcal{L}_m = \frac{r-l+b-t}{w + h}
\end{aligned}
$}
\end{gather}
We use a linear combination of the width and height of the rectangular region to avoid bias in the update of edges. Otherwise, if we use the ratio of area ($i.e.,$ $(r_{\!}-_{\!}l)_{\!}\times_{\!}(b_{\!}-_{\!}t)/(w\times h)$) as the mask loss, parameters of the longer edge ($e.g.,$ $b$ and $t$ when $(b-t)_{\!}<_{\!}(r-l)$ ) would have larger gradients and tend to change faster than the shorter edges, which leads to a bias towards updating the longer-edge parameters. Using a linear combination avoids this problem and each mask parameter has the same weight. 

Although we define a rectangular patch region, the final patch mask can be an arbitrary shape within the region. As shown in Fig.~\ref{fig:mask_def}c, given a predefined patch shape mask $m_s$ ($m_s[i, j] \in \{0, 1\}$), the final patch mask $m_p'^{\Theta}$ is calculated by element-wise multiplying the scaled shape mask $m'_s$  with the region mask $m^{\Theta}_p$ inside the rectangular region. Specifically, in each iteration, given border parameters $\Theta$, we can scale and fit the predefined shape mask $m_s$ into the center of the rectangular region getting mask $m'_s$ , which is denoted by the red color in Fig.~\ref{fig:mask_def}c. The final patch mask is calculated with Equation~\ref{eq:mask_shape} by multiplying the region mask and the shape mask within the rectangular region. Without loss of generality, we focus on rectangular shapes ($i.e.,$ $m_s \equiv 1$) in our evaluation.
\vspace{-5pt}
\begin{gather}\label{eq:mask_shape}
\scalebox{\eqscale}{$
\begin{aligned}
    m'^{\Theta}_p[i, j] =\left\{ \begin{matrix}
	m^{\Theta}_p[i, j] * m'_s[i, j] & i \in l...r, j \in t...b \\
	m^{\Theta}_p[i, j] & others
	\end{matrix}  \right.
\end{aligned}
$}
\end{gather}

In addition, our mask definition also supports optimizing with multiple patches. The key point is to take the union of several regions and optimize them together. We leave the details in Appendix~\ref{sec:appx_multi} and focus on one patch in our main results.

\vspace{-10pt}
\subsection{Attack Camouflage}\label{sec:atk_cam}

Patches generated in existing adversarial attacks against depth estimation models have obvious perturbations as shown in Fig.~\ref{fig:motivation}a. Unlike them, we use style transfer to camouflage the attack with natural styles. There have been works using style transfer~\cite{duan2020adversarial} in attacking classification models but we are the first to combine style transfer with the more challenging depth estimation attack. We use deep photo style transfer~\cite{luan2017deep} as our style transfer method. This method is a kind of neural style transfer which has demonstrated remarkable results for image stylization~\cite{gatys2016image}. It uses a convolutional neural network (CNN) to extract the deep features of an image and separate the content and style information in the deep feature representations. The source image will be updated iteratively to approach the style information extracted from the style image and keep the content information of the source image. Specifically, as defined in deep photo style transfer~\cite{luan2017deep}, there are four terms regarding the style transfer components in the loss function. They are style loss ($\mathcal{L}_s$), content loss ($\mathcal{L}_c$), smoothness loss ($\mathcal{L}_t$) and photorealism regularization loss ($\mathcal{L}_r$). The definition of these four style transfer losses can be found in Appendix~\ref{append:style_losses} and we refer the readers to~\cite{luan2017deep} for more detailed explanation on each term. The style transfer loss is the sum of those as follows:
\vspace{-10pt}
\begin{gather}
\scalebox{\eqscale}{$
\begin{aligned}
    \mathcal{L}_{st} = \mathcal{L}_s + \mathcal{L}_c + \mathcal{L}_t + \mathcal{L}_r
\end{aligned}
$}
\end{gather}
\vspace{-15pt}

In summary, our adversarial patch generation process can be formulated by the following optimization problem:
\vspace{-5pt}
\begin{gather}\label{eq:overall_optimization}
\scalebox{\eqscale}{$
\begin{aligned}
    \min\limits_{x', \Theta} ~~& \mathcal{L}_a  + \mathcal{L}_m + \lambda\mathcal{L}_{st}\\
    s.t.~~& x' \in [0, 255]^{3 \times w\times h}, 
    \Theta = \{l, r, t, b\}\\&
    0\leq l \leq r \leq w,~0 \leq t \leq b \leq h,
\end{aligned}
$}
\end{gather}
where $\lambda$ is an adjustable weight parameter to balance the style transfer naturalness and attack performance. The ablation study on different values of $\lambda$ can be found in Appendix~\ref{append:ablation}. The weights of other terms are fixed in our experiments. In each iteration, we calculate gradients of $x'$ and $\Theta$ with back propagation and, same as in deep photo style transfer~\cite{luan2017deep}, we use LBFGS~\cite{byrd1995limited} to update the patch $x'$. We update border parameters $\Theta$ with Adam~\cite{kingma2014adam} and we only update the edge with the maximum absolute gradient instead of four, which avoids the constraint of compressing the region from all directions in each iteration and provides more flexibility. We set a target ratio of the patch region in advance ($i.e.,$ the area of the patch region relative to the object) as the stopping criteria of mask optimization. In other words, the mask will stop updating when it is smaller than the predefined target ratio.
\vspace{-10pt}
\section{Experiments}
\vspace{-5pt}
\subsection{Experimental Setup}
\begin{table}[t]
    \centering
    \begin{minipage}{0.25\textwidth}
    \includegraphics[width=\textwidth]{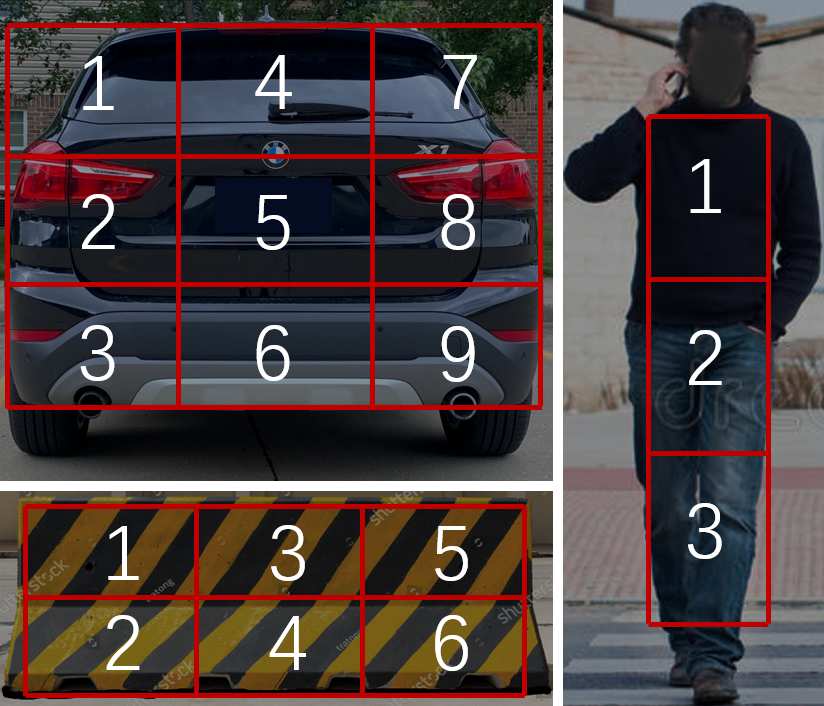}
    \captionof{figure}{\small Target objects and fixed regions.}
    \label{fig:objs_split}
    \vspace{-5pt}
    \includegraphics[width=\textwidth]{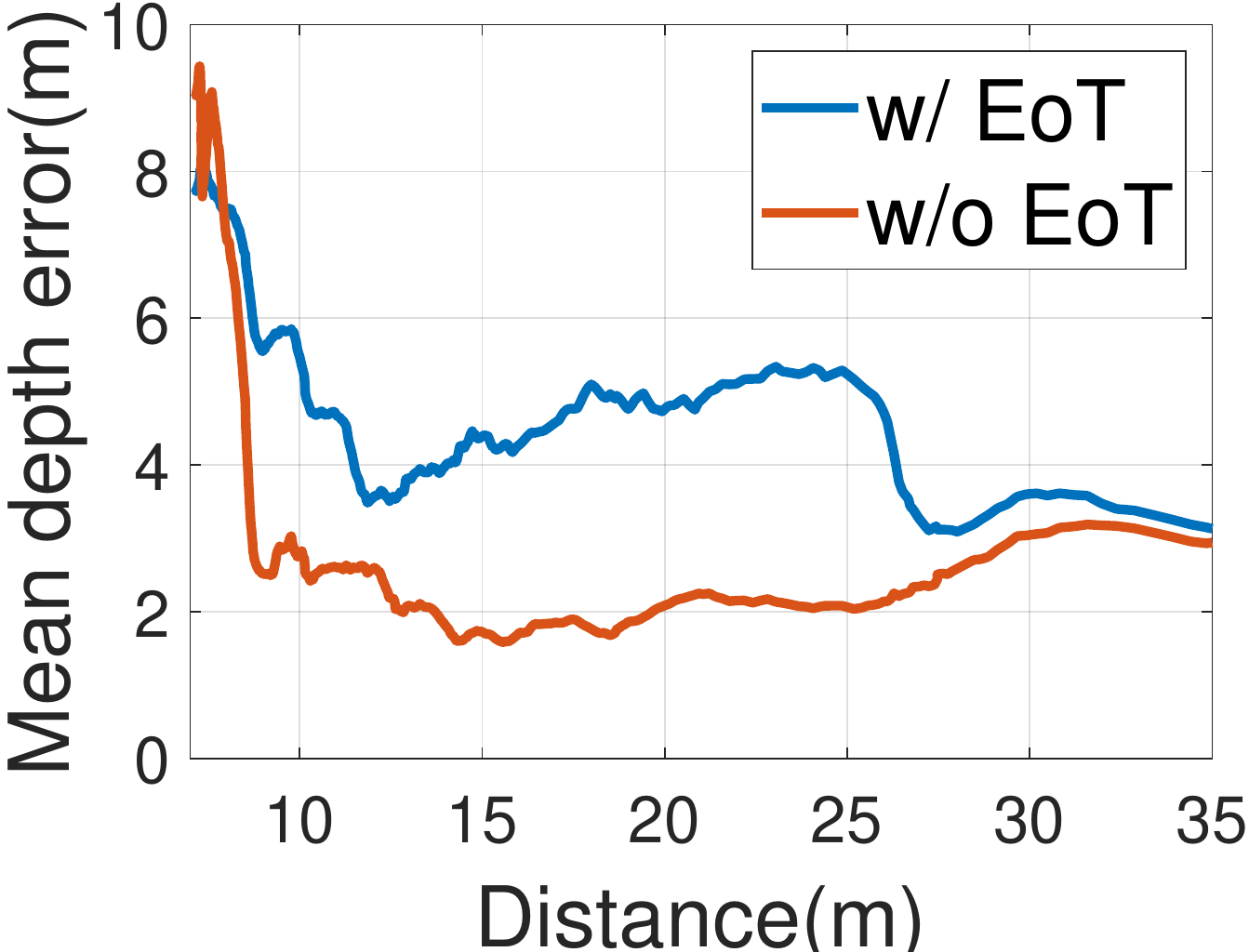}
    \captionof{figure}{\small $\mathcal{E}_d$ of the target vehicle at \textbf{different distance}.}
    \label{fig:attack_robust}
    \end{minipage}
    \hspace{10pt}
    \begin{minipage}{0.6\textwidth}
    \footnotesize
\vspace{-20pt}
\caption{\small Mean depth estimation error ($\mathcal{E}_d$) in attacking \textbf{fixed regions and optimized regions}.}
\vspace{5pt}
\scalebox{0.73} {
\begin{threeparttable}
\begin{tabular}{cccccccccc}
\toprule
\rowcolor{mygray}
 & \multicolumn{3}{c|}{Mono} & \multicolumn{3}{c|}{DH} & \multicolumn{3}{c}{Many} \\ 
\rowcolor{mygray}
 & V & TB & P & V & TB & P & V & TB & P \\ \midrule
Ours & \textbf{16.84} & \textbf{8.26} & \textbf{14.06} & \textbf{15.23} & \textbf{4.54} & \textbf{13.17} &\textbf{ 6.31} & \textbf{3.57} & \textbf{10.15} \\ \midrule
LO & 13.90 & 5.21 & 11.53 & 2.51 & 1.63	& 10.79	& 3.03 & 2.94 & 8.93  \\ \midrule
R1 & 3.70 & 2.35 & 10.20 & 2.25 & 1.50 & 11.78 & 1.12 & 2.77 & 9.21 \\ \midrule
R2 & 7.41 & 2.67 & 11.28 & 4.66 & 1.40 & 10.52 & 4.23 & 1.40 & 8.66 \\ \midrule
R3 & 5.20 & 4.96 & 5.05 & 3.92 & 1.45 & 4.08 & 1.33 & 3.05 & 5.06 \\ \midrule
R4 & 7.31 & 1.59 & - & 5.58 & 1.59 & - & 4.89 & 1.59 & - \\ \midrule
R5 & 14.95 & 2.39 & - & 7.70 & 0.90 & - & 5.66 & 2.43 & - \\ \midrule
R6 & 9.69 & 2.59 & - & 2.37 & 0.49 & - & 1.36 & 1.15 & - \\ \midrule
R7 & 3.23 & - & - & 2.62 & - & - & 1.67 & - & - \\ \midrule
R8 & 7.74 & - & - & 4.44 & - & - & 4.91 & - & - \\ \midrule
R9 & 5.36 & - & - & 1.38 & - & - & 1.32 & - & - \\ \bottomrule
\end{tabular}
\begin{tablenotes}
	\item{\footnotesize Mono: Monodepth2, DH: DepthHints, Many: Manydepth}
	\item{\footnotesize V: Vehicle, TB: Traffic Barrier, P: Pedestrian}
	\item{\footnotesize LO: Location Optimize in~\cite{rao2020adversarial}, R: Region}
\end{tablenotes}
\end{threeparttable}
}
\centering
\label{tab:effectiveness}
    \end{minipage}
\vspace{-20pt}
\end{table}

\noindent\textbf{MDE Model Selection.} In our evaluation, we use three monocular depth estimation models: Monodepth2~\cite{monodepth2}, Depthhints~\cite{watson-2019-depth-hints}, and Manydepth~\cite{watson2021temporal-manydepth}. We selected these models considering representativeness, practicality and open models. The details of the model selection criteria can be found in Appendix~\ref{append:model_selection}. 

\noindent\textbf{Target Object Selection.} Our attack is generic so it can be applied to any class of objects on public roads. This paper focuses on three representative types of objects to attack: vehicles, traffic barriers, and pedestrians as shown in Fig.~\ref{fig:objs_split}. We choose them because they are most common on public roads in regular driving scenarios, and a failure in detecting them could lead to life-threatening consequences. Vehicles are the most attractive objects for attackers since they are the main targets of perception systems on autonomous driving cars. We mainly focus on vehicles in our experiments.

\noindent\textbf{Evaluation Scene Selection.} We select 100 real-world driving scenes from KITTI dataset~\cite{Geiger2012CVPR-kittidataset} to evaluate the attack performance of the generated patch on each object. These scenes cover a wide range of roads ($e.g.,$ high-way, local, and rural roads) and background objects ($e.g.,$ trucks, traffic lights, and cars).

\noindent\textbf{Evaluation Metrics.} We use mean depth estimation error ($\mathcal{E}_d$) of the target object and ratio of affected region ($\mathcal{R}_a$) as our evaluation metrics. We use depth estimation of the original object as the ground truth and compare with depth estimation of the adversarial object. The mean depth estimation error denotes the attack effectiveness of our adversarial patch. The larger it is, the better the performance. Equation~\ref{eq:depth_metrics_diff} is the formal definition. Meanings of the symbols are the same as those in \S\ref{sec:method}.
\vspace{-5pt}
\begin{gather}\label{eq:depth_metrics_diff}
\scalebox{\eqscale}{$
\begin{aligned}
   \mathcal{E}_d = \frac{\text{sum}\left(|\mathcal{D}\left(\Lambda(O, R)\right) - \mathcal{D}\left(\Lambda(O',R)\right)|\odot M_o\right)}{\text{sum}(M_o)}
\end{aligned}
$}
\end{gather}
The ratio of affected region ($\mathcal{R}_a$) is defined in Equation~\ref{eq:depth_metrics_area}, where $\mathbf{I}(x)$ is the indicator function that evaluates to 1 only when $x$ is true. We define $\geq$ 10 meters error of depth estimation for a pixel as a valid attack and this pixel will be included in the affected region. $\mathcal{R}_a$ is the ratio between the number of affected pixels and all pixels of the object. A large value indicates that a broad area is affected. 
\vspace{-15pt}
\begin{gather}\label{eq:depth_metrics_area}
\scalebox{\eqscale}{$
\begin{aligned}
   \mathcal{R}_a = \frac{\text{sum}\left(\mathbf{I}\left(|\mathcal{D}\left(\Lambda(O, R)\right) - \mathcal{D}\left(\Lambda(O',R)\right)|\odot M_o\geq 10\right)\right)}{\text{sum}(M_o)}
\end{aligned}
$}
\end{gather}

\begin{figure}[t]
\vspace{-20pt}
    \centering
    \begin{minipage}{0.45\textwidth}
        \begin{figure}[H]
        	\includegraphics[width=\textwidth]{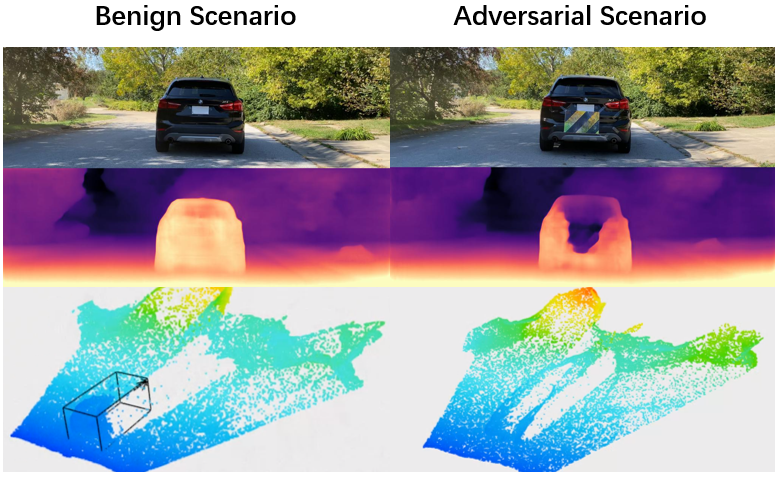}
        	\centering
        	\caption{\textbf{Physical world attack} example.(Video: \url{https://youtu.be/L-SyoAsAM0Y})}
        	\label{fig:physical_perform}
        \end{figure}
    \end{minipage}
    \begin{minipage}{0.53\textwidth}
        \begin{table}[H]
        \footnotesize
        \vspace{-30pt}
        \caption{\small \textbf{Physical world attack} result.}
        \scalebox{0.7} {
        \begin{tabular}{cccccc}
        \toprule
        \rowcolor{mygray}
         & Time (s) & Frames & $\mathcal{E}_{d}$ & Detected & Detection Rate \\ \midrule
        Route 1 Benign & 95 & 477 & 0.52 &469 & 98.32\% \\ \midrule
        Route 2 Benign & 82 & 412 & 0.77 &354 & 85.92\% \\ \midrule
        Route 3 Benign & 80 & 402 & 0.62 &348 & 86.57\% \\ \midrule\midrule
        \textbf{Total Benign} & \textbf{257} & \textbf{1291} & \textbf{0.64 }& \textbf{1171} & \textbf{90.70\%} \\ \midrule\midrule
        Route 1 Adv. & 94 & 468 & 6.73 & 45 & 9.62\% \\ \midrule
        Route 2 Adv. & 82 & 408 & 8.92 & 11 & 2.70\% \\ \midrule
        Route 3 Adv. & 80 & 402 & 7.68 & 10 & 2.49\% \\ \midrule\midrule
        \textbf{Total Adv.} & \textbf{256} & \textbf{1278} & \textbf{7.77} & \textbf{66} & \textbf{5.16\%} \\ \bottomrule
        \end{tabular}
        }
        \centering
        \label{tab:physical_obj_detect}
        \end{table}
    \end{minipage}
    \vspace{-20pt}
\end{figure}

\vspace{-15pt}
\subsection{Main results}
\vspace{-5pt}
We present our main results regarding effectiveness, robustness and stealth.

\noindent\textbf{Attack Effectiveness.} We run our attack with the three MDE models and we target the three types of objects for each model. For each object, we split it into several regions with equal size as shown in Fig.~\ref{fig:objs_split} and attack these fixed regions respectively ($i.e.,$ optimize the patch on each region.), then we compare with two patch region optimization techniques: our sensitive region localization (\S\ref{sec:sen_loc}) and the location-optimized patch~\cite{rao2020adversarial}. In~\cite{rao2020adversarial}, the authors update the location of a fixed-size patch after each optimization iteration. They tentatively move the patch towards four directions with a predefined stride and select the direction with the least adversarial loss as the next patch location.
For a fair comparison, we set the target ratio of patch region the same as that of those fixed regions ($e.g.$, 1/9 of the vehicle's read area). Our regional optimization stops when the patch ratio is smaller than the target ratio. In each test, we evaluate the mean depth estimation error ($\mathcal{E}_d$) of the target object in 100$_{\!}$ scenes and take the average of them as the result. In each scene, the object is placed at 7~m$_{\!}$ away from the victim's camera. We choose 7~m$_{\!}$ since it is the breaking distance~\cite{break_distance} while driving at a speed of 25~mph, which is almost the lowest in normal driving. In other words, it is the smallest distance at which the object has to be detected by the victim to avoid a crash in normal driving scenarios~\cite{cao2021invisible}. For the physical world experiments, our experimental setup can be found in Appendix~\ref{append:phy_setting}.

Table~\ref{tab:effectiveness} reports the effectiveness evaluation result. 
As shown, our attack is generic and effective on different depth estimation models and objects. With our sensitive region localization, an adversarial patch with 1/9 of the vehicle's rear area causes at least  6~m $\mathcal{E}_{d}$ across different depth estimation models. 
Observe that attack performance differs with patch regions. Our sensitive region localization can locate an optimal place that outperforms all those fixed regions and the location optimized regions in~\cite{rao2020adversarial}. For the physical world experiments, Fig.~\ref{fig:physical_perform} presents an example. As shown, the adversarial patch on the vehicle fools the vehicle's depth estimation, and the effect is not limited to the patch area but propagates to a broader area. After being projected to 3D space, it is more obvious that the point cloud of the adversarial vehicle is distorted comparing with the benign one. Table~\ref{tab:physical_obj_detect} reports the physical world attack performance. The first column in the table denotes different drives. The second column shows the time of each drive in seconds. The third column shows the total frames evaluated from the video, and we evaluate frames at a frequency of 5~Hz. The fourth column reports the mean depth estimation error ($\mathcal{E}_{d}$) of the vehicle. As shown, in benign scenarios, the error is under 1~m while the error in adversarial scenarios is over 7~m, which justifies the effectiveness of our attack in the physical world.

\begin{figure}[t]
\vspace{-20pt}
    \centering
    \begin{minipage}{0.38\textwidth}
        \begin{figure}[H]
            \begin{subfigure}[t]{0.49\columnwidth}
                \centering
                \includegraphics[width=\columnwidth]{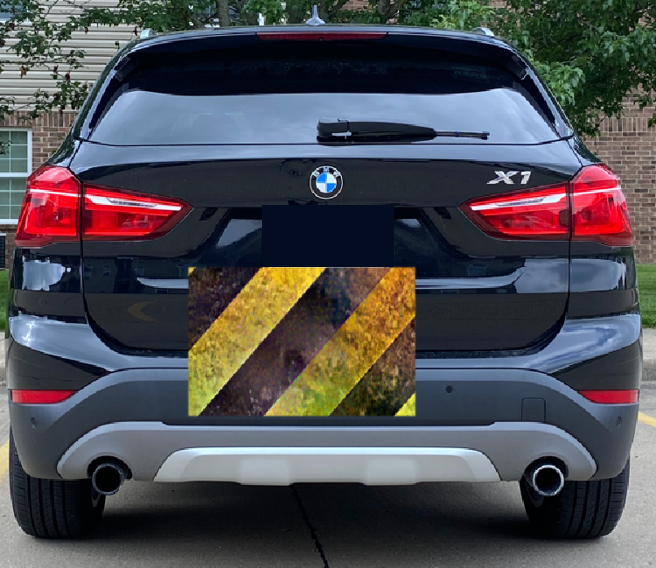}
                \caption{\small Ours }
                \label{fig:stealthy_ours}
            \end{subfigure}    
            \begin{subfigure}[t]{0.49\columnwidth}
                \centering
                \includegraphics[width=\columnwidth]{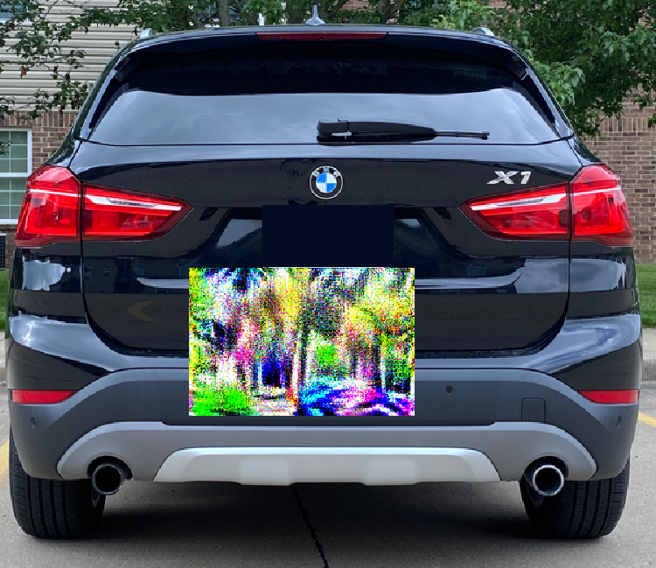}
                \caption{\small Baseline }
                \label{fig:stealthy_baseline}
            \end{subfigure}   
        \caption{\small \textbf{Naturalness} comparison.}\label{fig:stealthy_eval}
        \end{figure}
    \end{minipage}
    \begin{minipage}{0.6\textwidth}
        \begin{figure}[H]
            \begin{subfigure}[t]{.49\columnwidth}
        		\centering
        		\includegraphics[width=\columnwidth]{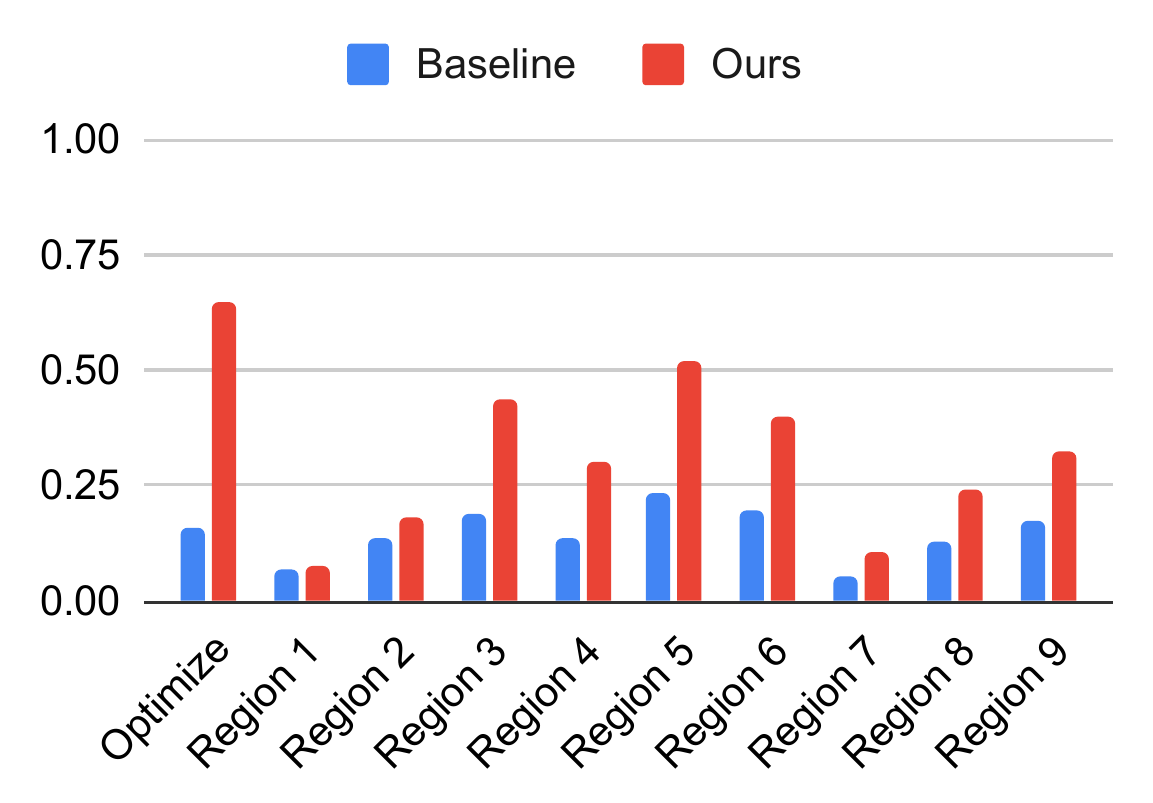}
        		\vspace{-15pt}
        		\caption{\small $\mathcal{R}_a$}
        		\label{fig:baseline_ratio}
        	\end{subfigure}    
        	\begin{subfigure}[t]{.49\columnwidth}
        		\centering
        		\includegraphics[width=\columnwidth]{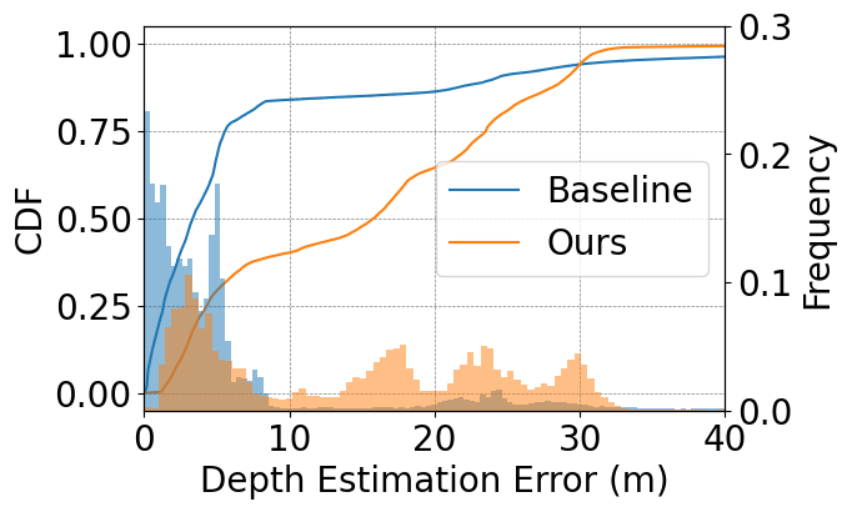}
        		\vspace{-15pt}
        		\caption{\small CDF}
        		\label{fig:cdf_baseline}
        	\end{subfigure}    
        	\vspace{-10pt}
            \caption{\small Comparing \textbf{patch-oriented attack} (baseline) with our \textbf{object-oriented attack}.}\label{fig:stealthy_perform}
        \end{figure}
    \end{minipage}
    \vspace{-20pt}
\end{figure}

\noindent\textbf{Attack Robustness.} Relative to the victim vehicle, we place the adversarial object at places with longitudinal distances$_{\!}$ ($i.e.,$$_{\!}$ forward and back)$_{\!}$ ranging from 7~m$_{\!}$ to 35~m$_{\!}$ and lateral distances$_{\!}$ ($i.e.,$$_{\!}$ left and right)$_{\!}$ ranging from -1~m$_{\!}$ to 1~m. The $_{\!}$ 7~m$_{\!}$ to 35~m$_{\!}$ longitudinal distance corresponds to the brake distance for driving speed from about 25$_{\!}$ to 55$_{\!}$ mph~\cite{brake_speed}.$_{\!}$ We$_{\!}$ consider$_{\!}$ the victim$_{\!}$ vehicle$_{\!}$ at the center of the lane, and$_{\!}$ -1~m$_{\!}$ to$_{\!}$ 1~m of lateral$_{\!}$ deviation from$_{\!}$ the lane$_{\!}$ center$_{\!}$ covers$_{\!}$ most$_{\!}$ driving scenarios of the vehicle ahead~\cite{dominguez2016comparison}. We use a vehicle as the target object and Monodepth2 as the depth estimation network. We use the regional optimization and set the target patch size to 1/9 of the vehicle's rear area. We test our attack with and without EoT~\cite{athalye2018synthesizing} (see \S\ref{sec:adv_gen})
during optimization. 

Fig.~\ref{fig:attack_robust} shows the result of the robustness evaluation. We report the mean depth estimation error of the target object under different longitude distances with the victim vehicle. Observe that our attack is robust and causes more than 3~m of mean depth estimation error in different victim approaching positions. EoT increases the attack performance by 40.63\% and makes our attack more robust in different distances.  
As shown, the closer the target object, the larger the error in depth estimation, which makes the victim vehicle harder to detect the object from the distorted pseudo-Lidar and continue approaching it until collision. In the physical world experiments, our attack is conducted with real driving scenarios. Compared to evaluating with a single image from a specific position in prior work, continuous and dynamic movement is more challenging and practical. Our attack is shown to be robust under different lighting conditions ($e.g.,$ shadows and different light directions), driving operations ($e.g.,$ moving straight and turning) and background scenes. The dynamic moving video of our physical world attack is at \url{https://youtu.be/L-SyoAsAM0Y}.

\noindent\textbf{Stealth} As we discussed in our motivation, we consider the stealth in two directions: the naturalness of appearance and the patch size. In terms of naturalness, we compare the adversarial patch generated by our method with the baseline method proposed by Yamanaka et al.~\cite{yamanaka2020adversarial}. As shown in Fig.~\ref{fig:stealthy_eval}, our method with style transfer-based camouflage generates more natural patches and is less likely to be identified as adversarial but just a normal sticker. Human studies conducted in~\cite{luan2017deep,duan2020adversarial} also justify the naturalness of style-transfer-based image processing. As for the patch size, a smaller size suggests more stealth and less effectiveness. We hence investigate maximizing the attack effect with small patches. We compare the $\mathcal{R}_a$ caused by our object-oriented attack and the patch-oriented attack in~\cite{yamanaka2020adversarial} which only attacks the patch area in their adversarial loss design instead of considering the whole object. For a fair comparison, we use style-transfer-based camouflage in both methods and we test with fixed regions and the regional optimization. This experiment is conducted on Monodepth2~\cite{monodepth2} targeting the vehicle and other settings are the same as the previous setup in effectiveness evaluation.

As shown in Fig.~\ref{fig:baseline_ratio}, our method (object-oriented) has over 2.5 times higher $\mathcal{R}_a$ on the vehicle than the baseline (patch-oriented) in all cases, and our method in the regional optimization case outperforms all other fixed-region cases. Hence, with the same total patch area, our object-oriented attack with regional optimization affects a broader area than the baseline. In other words, \textit{to achieve similar attack effect, using our method requires a smaller patch and is thus stealthier.} Fig.~\ref{fig:cdf_baseline} additionally shows the CDF and histogram of depth estimation error in the case with our regional optimization. As shown, more than 80\% errors caused by the baseline method are below 10~m, which corresponds to our observation in Fig.~\ref{fig:motivation}c that the patch-oriented attack mainly affects the limited patch area and the effect of our method propagates to a broader area causing larger errors.

Evaluations on the transferability of our attack can be found in Appendix~\ref{append:transfer}.

\begin{figure}[t]
    \centering
    \vspace{-15pt}
    \begin{minipage}{0.38\textwidth}
        \begin{figure}[H]
            \begin{subfigure}[t]{0.49\columnwidth}
                \centering
                \includegraphics[width=\columnwidth]{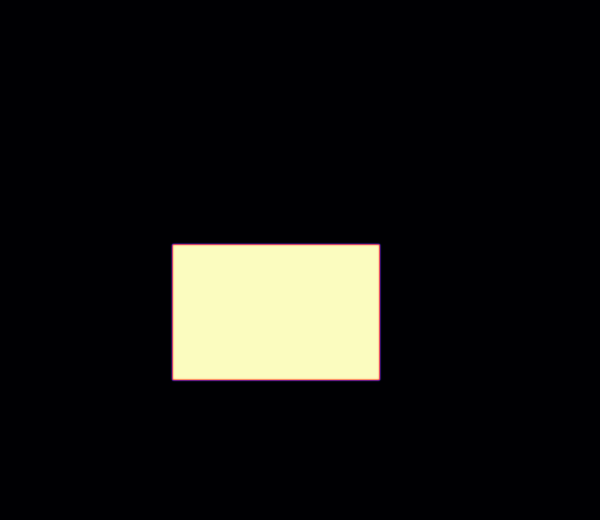}
                \vspace{-4pt}
                \caption{\small Regional }
                \label{fig:depth_baseline_mask_a}
            \end{subfigure}    
            \begin{subfigure}[t]{0.49\columnwidth}
                \centering
                \includegraphics[width=\columnwidth]{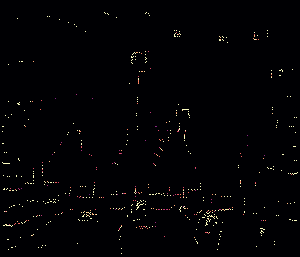}
                \vspace{-4pt}
                \caption{\small Per-pixel }
                \label{fig:depth_baseline_mask_b}
            \end{subfigure}   
            \vspace{-10pt}
        \caption{\small Different \textbf{mask optimization methods}.}\label{fig:depth_baseline_mask}
        \end{figure}
    \end{minipage}
    \begin{minipage}{0.6\textwidth}
        \begin{figure}[H]
            \begin{subfigure}[t]{.49\columnwidth}
        		\centering
        		\includegraphics[width=\columnwidth]{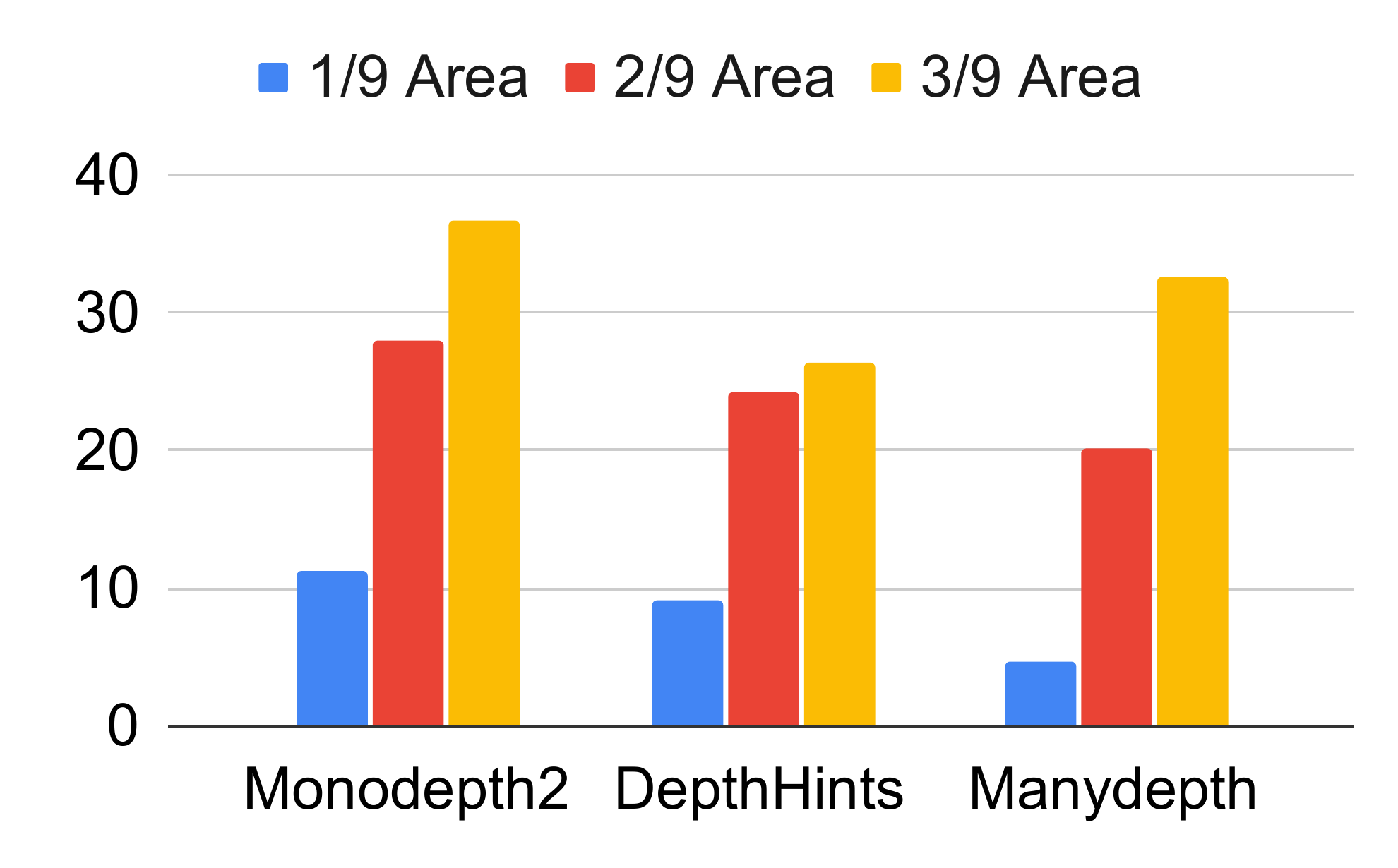}
        		\caption{$\mathcal{E}_d$ }
        		\label{fig:stealth_size_a}
        	\end{subfigure}    
        	\begin{subfigure}[t]{.49\columnwidth}
        		\centering
        		\includegraphics[width=\columnwidth]{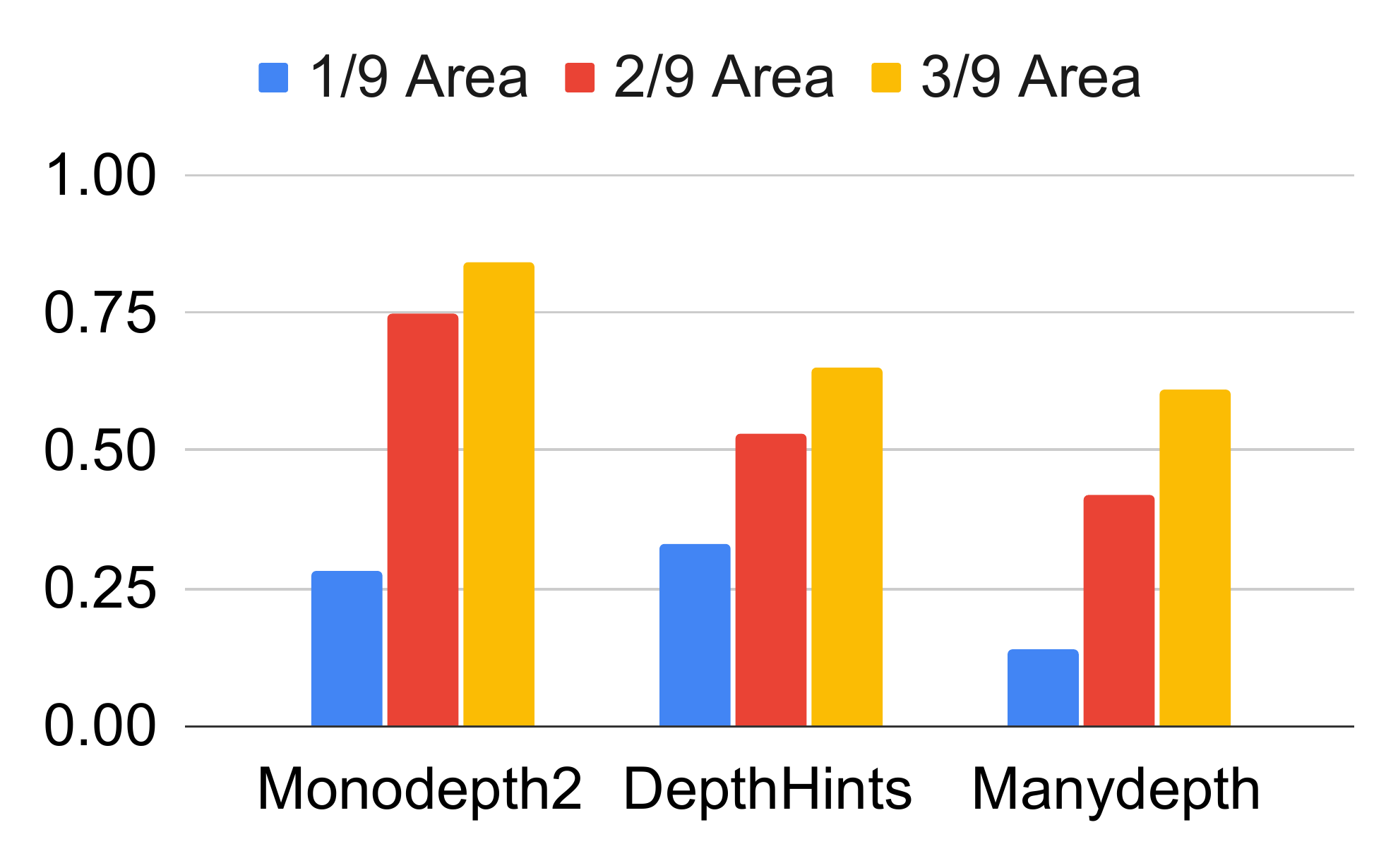}
        		\caption{$\mathcal{R}_a$ }
        		\label{fig:stealth_size_b}
        	\end{subfigure}    
        	\vspace{-10pt}
            \caption{\small Attack performance of regional optimization with \textbf{different target sizes}.}\label{fig:stealth_size}
        \end{figure}
    \end{minipage}
    \vspace{-15pt}
\end{figure}

\vspace{-10pt}
\subsection{Ablation Study}
\vspace{-5pt}
We investigate our method through an ablation study. 
\setlength{\intextsep}{0pt}
\begin{wraptable}{R}{4cm}
    \centering
    \caption{\small \textbf{Ablation study
    }.}\label{tab:ablation}
    \vspace{-10pt}
    \scalebox{0.7}{
    \begin{threeparttable}
    \begin{tabular}{p{1.3cm} p{1.3cm} p{1.3cm} p{1.3cm}}
    \toprule
    \rowcolor{mygray}
    OA & RO & $\mathcal{E}_d$ & $\mathcal{R}_a$ \\
    \midrule
     &  & 8.47 & 0.23 \\\midrule
     & \checkmark & 6.38 & 0.16 \\\midrule
    \checkmark &  & 14.95 & 0.52 \\\midrule
    \checkmark & \checkmark & \textbf{16.84} & \textbf{0.65}\\
    \bottomrule
    \end{tabular}
    \begin{tablenotes}
    	\item OA: Object-oriented Adv. Loss
    	\item RO: Regional Optimization
    \end{tablenotes}
    \end{threeparttable}
    }
\end{wraptable}

\noindent\textbf{Combinations.} As described in \S\ref{sec:method}, we use the object-oriented adversarial loss design and the regional optimization of the patch mask to maximize the attack effect with a small patch. We conduct ablations on these techniques to see how each component contributes. Table \ref{tab:ablation} shows the result. We attack Monodepth2 and use the vehicle as the target object and report $\mathcal{E}_d$ and $\mathcal{R}_a$. For those tests without regional optimization, we use \#5 fixed region because its attack performance is the best among all the fixed regions in previous evaluations. As shown, the object-oriented adversarial loss itself can improve the attack performance while the regional optimization cannot. The regional optimization is useful only when object-oriented adversarial loss is applied together. It makes sense that the regional optimization has to consider the whole object to find an optimal place regarding the target object. However, the patch-oriented design does not encode the global information so our regional optimization cannot converge to the most effective region.

\noindent\textbf{Mask Optimization Methods.}
We compare our regional optimization with another commonly used mask optimization technique which treats pixels of the patch mask $m_p$ as optimizable parameters instead of the four borders. This method has been used in many backdoor scanning works such as Neural Cleanse~\cite{wang2019neural-NC} and ABS~\cite{liu2019abs} to find a trigger that modifies a limited portion of image and causes misclassification. Fig.~\ref{fig:depth_baseline_mask} shows the comparison. Observe that the patch mask generated by the baseline method is more sparse and scattered. The patch unit is tiny. Compared with our method, it is not suitable as a physical world attack vector because it is hard to print and deploy these scattered tiny patches while our regional patch is more practical.

\noindent\textbf{Patch Sizes.} Larger patches have more effect on depth estimation but are less stealthy. We evaluate our attack on a vehicle object with three different target patch sizes and use three depth estimation models. Fig.~\ref{fig:stealth_size} shows the result. Observe that the mean depth estimation error $\mathcal{E}_d$ and the ratio of affected region $\mathcal{R}_a$ increase with the size of patch for all three target networks. 

More ablation studies on the style transfer weight $\lambda$ are in Appendix~\ref{append:ablation}.

\begin{figure}[t]
    \centering
    \begin{minipage}{0.55\textwidth}
        \begin{figure}[H]
            \begin{subfigure}[t]{0.49\columnwidth}
                \centering
                \includegraphics[width=\columnwidth]{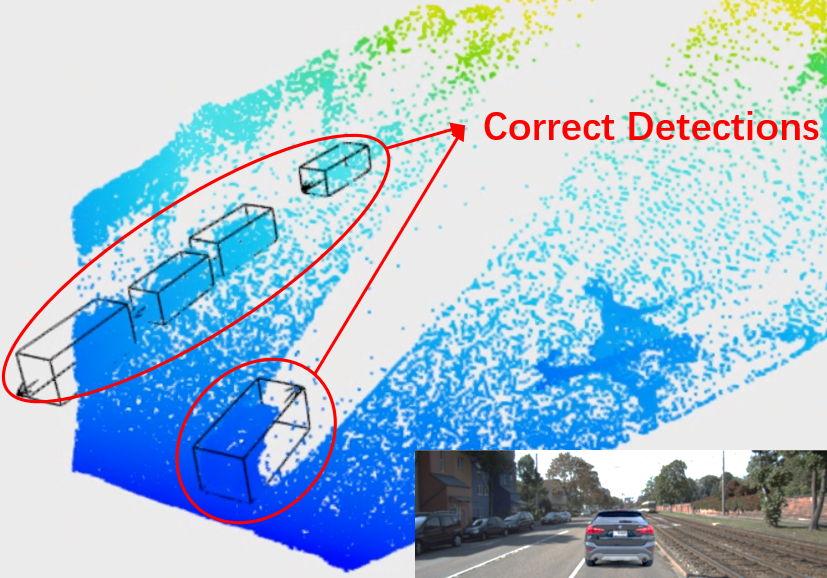}
                \caption{\small Benign }
		        \label{fig:cloud_ben_scene}
            \end{subfigure}    
            \begin{subfigure}[t]{0.49\columnwidth}
                \centering
                \includegraphics[width=\columnwidth]{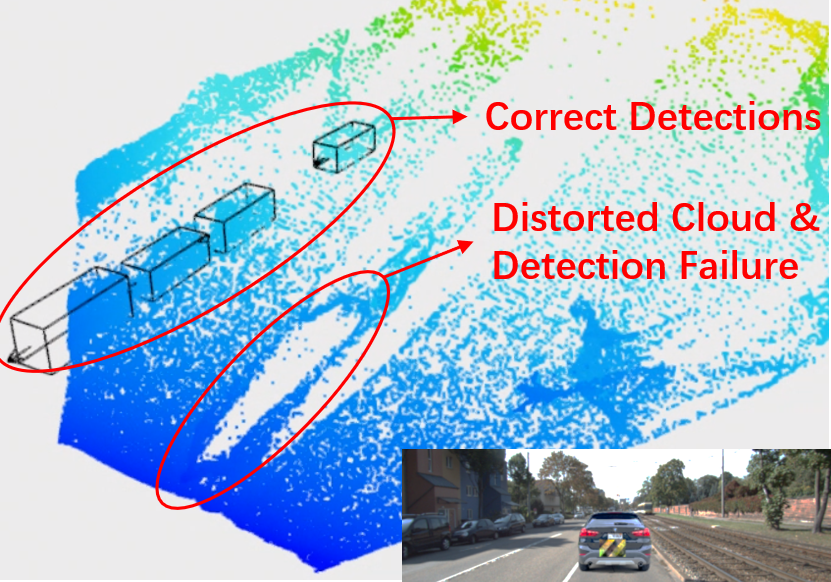}
                \caption{\small Adversarial }
		        \label{fig:cloud_adv_scene}
            \end{subfigure}   
        \vspace{-10pt}
        \caption{\small Attack against \textbf{3D object detection}.}\label{fig:obj_detect}
        \end{figure}
    \end{minipage}
    \begin{minipage}{0.4\textwidth}
        \begin{table}[H]
            \footnotesize
            \vspace{-32pt}
            \caption{\small \textbf{Attack success rate} of different adversarial patches.}
            \vspace{5pt}
            \scalebox{0.7} {
            \begin{tabular}{cccc}
            \toprule
            \rowcolor{mygray}
             & {Monodepth2} & {DepthHints} & {Manydepth} \\ 
            \midrule
            1/9 Area & 95\% & 93\% & 98\%      \\
            \midrule
            2/9 Area & 98\% & 97\% & 100\%     \\
            \midrule
            1/3 Area & 100\% & 100\% & 100\%   \\ 
            \bottomrule
            \end{tabular}
            }
            \centering
            \label{tab:obj_asr}
        \end{table}
    \end{minipage}
    \vspace{-15pt}
\end{figure}

\vspace{-10pt}
\subsection{Downstream Task Impact}\label{sec:downstream}
\vspace{-5pt}
We evaluate the impact of our attack on a point cloud based 3D object detection model -- PointPillars~\cite{lang2019pointpillars}
and use attack success rate (ASR) as the metric to evaluate our method on 3D object detection. We consider the attack is successful when the benign vehicle can be detected by PointPillar while the adversarial object cannot. Detailed setups can be found in Appendix~\ref{append:3D_model}.

Fig.~\ref{fig:obj_detect} gives an example of a successful attack. Fig.~\ref{fig:cloud_ben_scene} presents a benign scenario where the benign vehicle can be correctly detected with a 3D bounding box. Fig.~\ref{fig:cloud_adv_scene} shows the corresponding adversarial scenario where the pseudo-Lidar point cloud of the adversarial vehicle is severely distorted by the patch, and thus the vehicle is not detected. The PointPillar network can correctly detect the benign vehicle in all the 100 scenes and the attack success rate (ASR) of different adversarial patches are reported in Table~\ref{tab:obj_asr}. The first column denotes different patch sizes and columns 2-4 refer to the three different target networks. As shown, the ASR is over 90\% with all the patch sizes and target networks. Even when the patch size is just 1/9 of the vehicle's rear area, it can still achieve at least 93\% ASR, which shows that our attack is an effective method in fooling the 3D object detection model. In the physical world experiments, the  fifth column of Table~\ref{tab:physical_obj_detect} denotes the number of frames in which the vehicle is detected from the pseudo-Lidar point cloud, and the sixth column reports the object detection rate. For benign cases, the rate of successful object detection is 90.70\% in 1291 data frames. The rate drops to 5.16\% in adversarial cases with 1278 data frames, which shows that our attack is effective and degrades the object detection rate significantly.

\begin{figure*}[t]
\centering
	\begin{subfigure}[t]{.27\textwidth}
		\centering
		\includegraphics[width=\columnwidth]{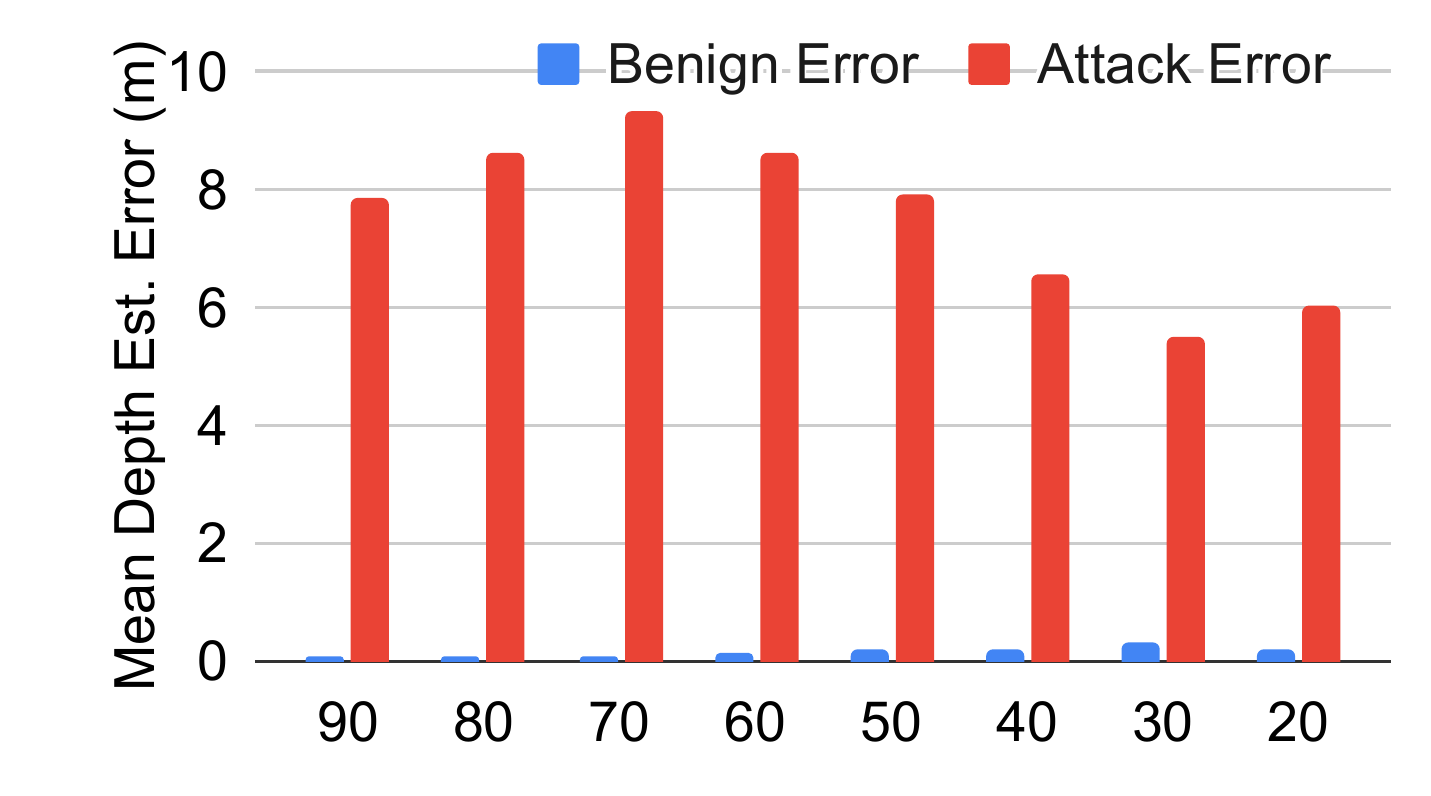}
		\caption{\scriptsize JPEG Compression }
		\label{fig:def_gpeg}
	\end{subfigure}    
	\begin{subfigure}[t]{.15\textwidth}
		\centering
		\includegraphics[width=\columnwidth]{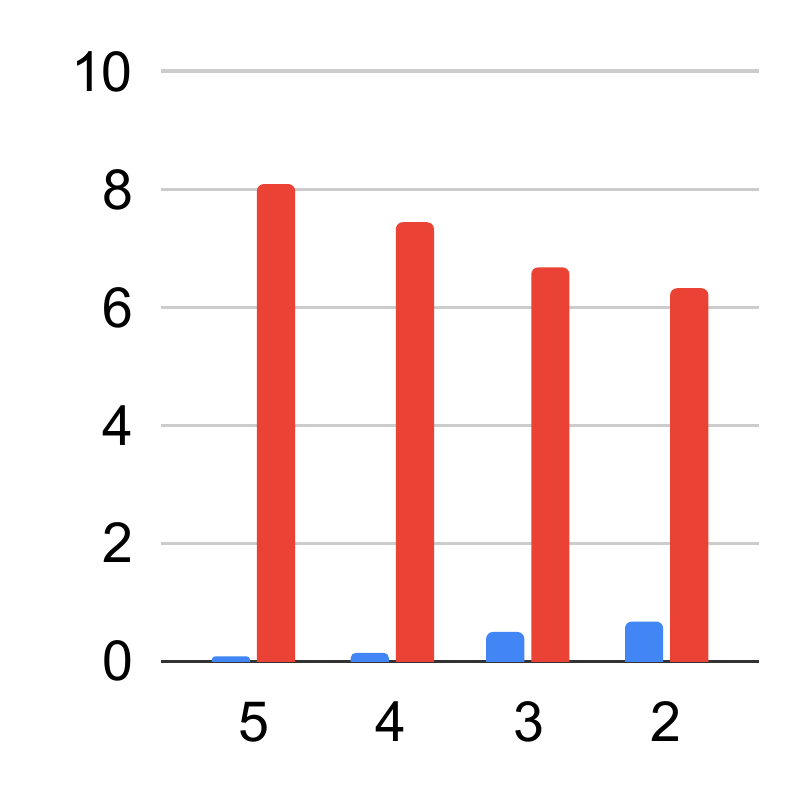}
		\caption{\scriptsize Bit-Depth }
		\label{fig:def_bit}
	\end{subfigure}   
	\begin{subfigure}[t]{.18\textwidth}
		\centering
		\includegraphics[width=\columnwidth]{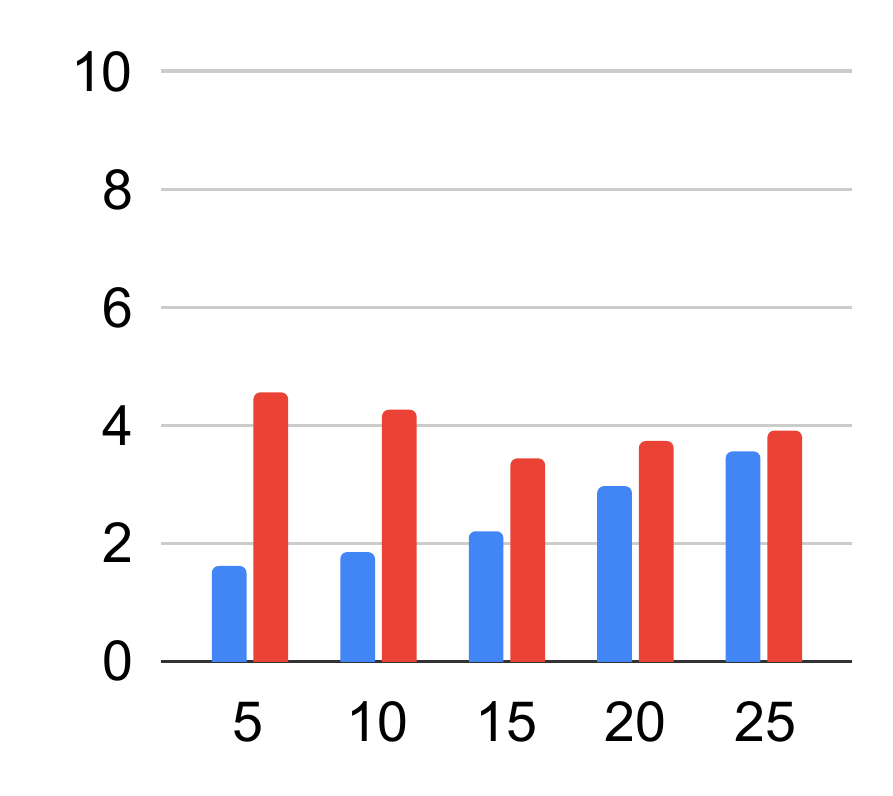}
		\caption{\scriptsize Median Blur }
		\label{fig:def_smoothing}
	\end{subfigure}   
    \begin{subfigure}[t]{.16\textwidth}
		\centering
		\includegraphics[width=\columnwidth]{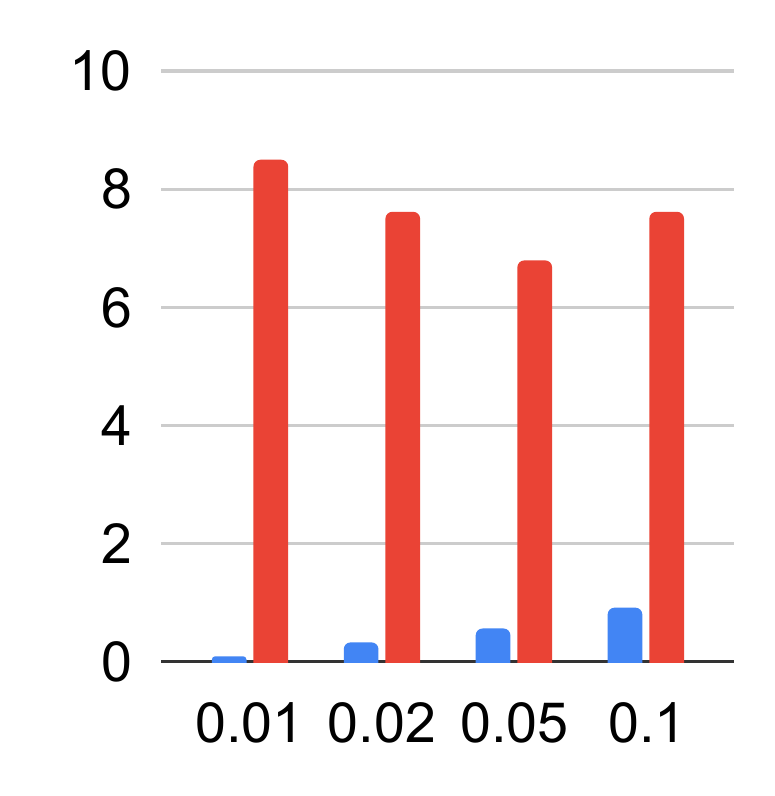}
		\caption{\scriptsize Noise }
		\label{fig:def_noise}
	\end{subfigure}   
	\begin{subfigure}[t]{.20\textwidth}
		\centering
		\includegraphics[width=\columnwidth]{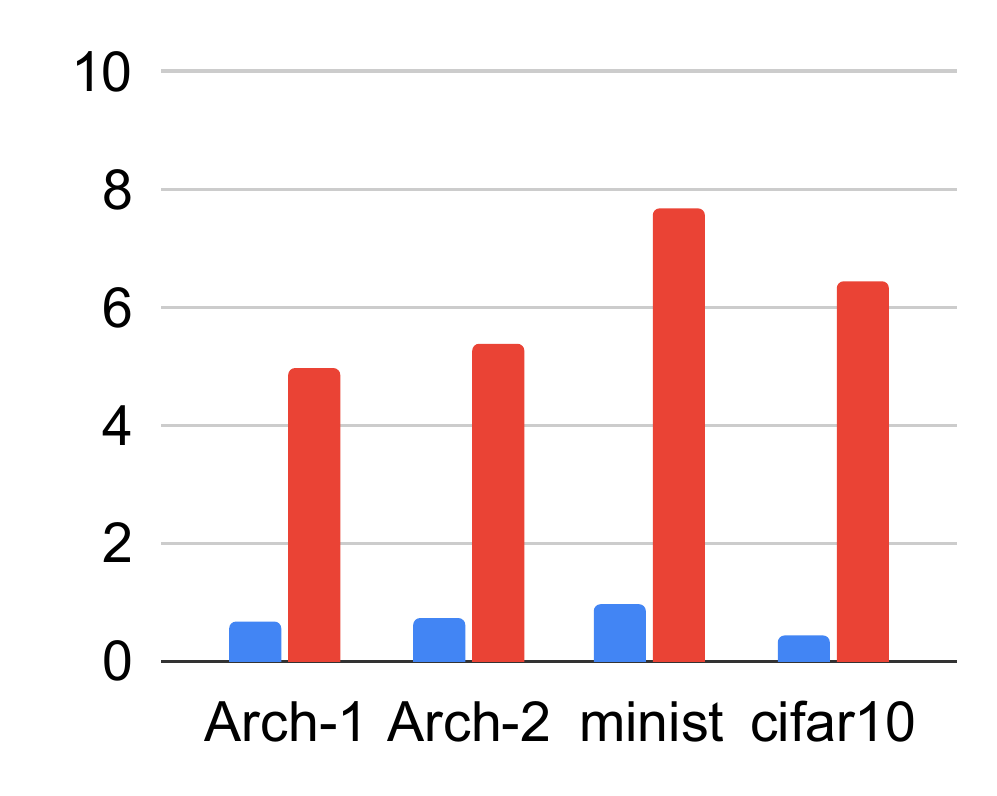}
		\caption{\scriptsize Autoencoder }
		\label{fig:def_encoder}
	\end{subfigure}   
	 \vspace{-8pt}
	\caption{\footnotesize Five directly-applicable defence methods. \textit{Benign Error}: Error caused by the defence in benign cases. \textit{Attack Error}: Error caused by our attack.}
	\label{fig:defence_methods}
	\vspace{-15pt}
\end{figure*}

\vspace{-10pt}
\subsection{Defence Discussion}\label{sec:defence}
\vspace{-5pt}
Although many defense techniques against adversarial examples have been proposed
, none of them focuses on MDE to the best of our knowledge.
As a best effort to understand the performance of our attack under different defences, we apply 5 popular defence techniques which perform input transformations without retraining the victim network. They are JPEG compression~\cite{dziugaite2016study}, bit-depth reduction~\cite{xu2017feature}, median blurring~\cite{xu2017feature}, adding Gaussian noise\cite{zhang2019defending} and autoencoder reformation~\cite{meng2017magnet}. Fig.~\ref{fig:defence_methods} presents our results. We report the$_{\!}$ $\mathcal{E}_d$$_{\!}$ of the benign vehicle and the adversarial vehicle under different input transformations. An ideal defence should minimize both errors.$_{\!}$ As shown,$_{\!}$ our attack can still cause over 5$_{\!}$ meters $\mathcal{E}_d$ in all methods except median blur.$_{\!}$ In median blur, the attack is mitigated but the benign performance also drops a lot. This shows that these techniques cannot effectively defend our attack without greatly harming the benign performance. 
This might be because these defenses are mainly for disrupting digital-space
human-imperceptible perturbations~\cite{sato2021dirty} instead of robust physical world attacks. 
Detailed descriptions and configurations of the above defences in our evaluation can be found in Appendix~\ref{append:def_methods}, as well as the discussion of other potential defences such as adversarial training and fusion-based methods.


\vspace{-10pt}
\section{Conclusion}
\vspace{-5pt}

We investigate stealthy physical-world adversarial patch attack against MDE in the AD scenario. We design a novel physical-object-oriented optimization framework to generate stealthy and effective adversarial patches via an object-oriented adversarial loss design, sensitive region localization and natural style based camouflage. Experimental results show that our attack is effective, stealthy and robust against different target objects, state-of-the-art models and a representative downstream task in AD. We achieve over 6 meters of mean depth estimation error for a real vehicle with a patch of 1/9 of the vehicle's rear area and have more than 90\% attack success rate in 3D object detection. Popular defence techniques using input transformations cannot defend our attack well.


\par\vfill\par

\clearpage
%
%
\bibliographystyle{splncs04}
\bibliography{bibliography}

\begin{thebibliography}{10}
\providecommand{\url}[1]{\texttt{#1}}
\providecommand{\urlprefix}{URL }
\providecommand{\doi}[1]{https://doi.org/#1}

\bibitem{baidu_apollo_lite}
{Baidu unveils Apollo Lite Level 4 vision-based autonomous driving solution},
  \url{https://autonews.gasgoo.com/m/Detail/70016068.html}

\bibitem{break_distance}
{Break Distance}, \url{http://www.csgnetwork.com/stopdistcalc.html}

\bibitem{depthhints_github}
{Depth Hints Github}, \url{https://github.com/nianticlabs/depth-hints}

\bibitem{manydepth_github}
{Manydepth Github}, \url{https://github.com/nianticlabs/manydepth}

\bibitem{monodepth2_github}
{Monodepth2 Github}, \url{https://github.com/nianticlabs/monodepth2}

\bibitem{tesla-AI-Day}
{Tesla AI day}, \url{https://youtu.be/j0z4FweCy4M?t=5295}

\bibitem{tesla-self-supervised}
{Tesla use per-pixel depth estimation with self-supervised learning},
  \url{https://youtu.be/hx7BXih7zx8?t=1334}

\bibitem{brake_speed}
{Vehicle Stopping Distance}, \url{http://www.csgnetwork.com/stopdistcalc.html}

\bibitem{aich2020bidirectional}
Aich, S., Vianney, J.M.U., Islam, M.A., Kaur, M., Liu, B.: Bidirectional
  attention network for monocular depth estimation. arXiv preprint
  arXiv:2009.00743  (2020)

\bibitem{athalye2018synthesizing}
Athalye, A., Engstrom, L., Ilyas, A., Kwok, K.: Synthesizing robust adversarial
  examples. In: International conference on machine learning. pp. 284--293.
  PMLR (2018)

\bibitem{brown2017adversarial}
Brown, T.B., Man{\'e}, D., Roy, A., Abadi, M., Gilmer, J.: Adversarial patch.
  arXiv preprint arXiv:1712.09665  (2017)

\bibitem{byrd1995limited}
Byrd, R.H., Lu, P., Nocedal, J., Zhu, C.: A limited memory algorithm for bound
  constrained optimization. SIAM Journal on scientific computing
  \textbf{16}(5),  1190--1208 (1995)

\bibitem{cao2021invisible}
Cao, Y., Wang, N., Xiao, C., Yang, D., Fang, J., Yang, R., Chen, Q.A., Liu, M.,
  Li, B.: Invisible for both camera and lidar: Security of multi-sensor fusion
  based perception in autonomous driving under physical-world attacks. In: 2021
  IEEE Symposium on Security and Privacy (SP). pp. 176--194. IEEE (2021)

\bibitem{cao2019adversarial}
Cao, Y., Xiao, C., Cyr, B., Zhou, Y., Park, W., Rampazzi, S., Chen, Q.A., Fu,
  K., Mao, Z.M.: Adversarial sensor attack on lidar-based perception in
  autonomous driving. In: Proceedings of the 2019 ACM SIGSAC conference on
  computer and communications security. pp. 2267--2281 (2019)

\bibitem{chen2018shapeshifter}
Chen, S.T., Cornelius, C., Martin, J., Chau, D.H.P.: Shapeshifter: Robust
  physical adversarial attack on faster r-cnn object detector. In: Joint
  European Conference on Machine Learning and Knowledge Discovery in Databases.
  pp. 52--68. Springer (2018)

\bibitem{dominguez2016comparison}
Dominguez, S., Ali, A., Garcia, G., Martinet, P.: Comparison of lateral
  controllers for autonomous vehicle: Experimental results. In: 2016 IEEE 19th
  International Conference on Intelligent Transportation Systems (ITSC). pp.
  1418--1423. IEEE (2016)

\bibitem{duan2020adversarial}
Duan, R., Ma, X., Wang, Y., Bailey, J., Qin, A.K., Yang, Y.: Adversarial
  camouflage: Hiding physical-world attacks with natural styles. In:
  Proceedings of the IEEE/CVF conference on computer vision and pattern
  recognition. pp. 1000--1008 (2020)

\bibitem{dziugaite2016study}
Dziugaite, G.K., Ghahramani, Z., Roy, D.M.: A study of the effect of jpg
  compression on adversarial images. arXiv preprint arXiv:1608.00853  (2016)

\bibitem{gatys2016image}
Gatys, L.A., Ecker, A.S., Bethge, M.: Image style transfer using convolutional
  neural networks. In: Proceedings of the IEEE conference on computer vision
  and pattern recognition. pp. 2414--2423 (2016)

\bibitem{Geiger2012CVPR-kittidataset}
Geiger, A., Lenz, P., Urtasun, R.: Are we ready for autonomous driving? the
  kitti vision benchmark suite. In: Conference on Computer Vision and Pattern
  Recognition (CVPR) (2012)

\bibitem{monodepth2}
Godard, C., {Mac Aodha}, O., Firman, M., Brostow, G.J.: Digging into
  self-supervised monocular depth prediction  (October 2019)

\bibitem{goodfellow2014explaining}
Goodfellow, I.J., Shlens, J., Szegedy, C.: Explaining and harnessing
  adversarial examples. arXiv preprint arXiv:1412.6572  (2014)

\bibitem{packnet}
Guizilini, V., Ambrus, R., Pillai, S., Raventos, A., Gaidon, A.: 3d packing for
  self-supervised monocular depth estimation. In: IEEE Conference on Computer
  Vision and Pattern Recognition (CVPR) (2020)

\bibitem{jia2020fooling}
Jia, Y., Lu, Y., Shen, J., Chen, Q.A., Zhong, Z., Wei, T.: Fooling detection
  alone is not enough: First adversarial attack against multiple object
  tracking. In: International Conference on Learning Representations (ICLR)
  (2020)

\bibitem{kingma2014adam}
Kingma, D.P., Ba, J.: Adam: A method for stochastic optimization. arXiv
  preprint arXiv:1412.6980  (2014)

\bibitem{komkov2021advhat}
Komkov, S., Petiushko, A.: Advhat: Real-world adversarial attack on arcface
  face id system. In: 2020 25th International Conference on Pattern Recognition
  (ICPR). pp. 819--826. IEEE (2021)

\bibitem{lang2019pointpillars}
Lang, A.H., Vora, S., Caesar, H., Zhou, L., Yang, J., Beijbom, O.:
  Pointpillars: Fast encoders for object detection from point clouds. In:
  Proceedings of the IEEE/CVF Conference on Computer Vision and Pattern
  Recognition. pp. 12697--12705 (2019)

\bibitem{lee2021bbam}
Lee, J., Yi, J., Shin, C., Yoon, S.: Bbam: Bounding box attribution map for
  weakly supervised semantic and instance segmentation. In: Proceedings of the
  IEEE/CVF Conference on Computer Vision and Pattern Recognition. pp.
  2643--2652 (2021)

\bibitem{levin2007closed}
Levin, A., Lischinski, D., Weiss, Y.: A closed-form solution to natural image
  matting. IEEE transactions on pattern analysis and machine intelligence
  \textbf{30}(2),  228--242 (2007)

\bibitem{liu2019abs}
Liu, Y., Lee, W.C., Tao, G., Ma, S., Aafer, Y., Zhang, X.: Abs: Scanning neural
  networks for back-doors by artificial brain stimulation. In: Proceedings of
  the 2019 ACM SIGSAC Conference on Computer and Communications Security. pp.
  1265--1282 (2019)

\bibitem{luan2017deep}
Luan, F., Paris, S., Shechtman, E., Bala, K.: Deep photo style transfer. In:
  Proceedings of the IEEE conference on computer vision and pattern
  recognition. pp. 4990--4998 (2017)

\bibitem{madry2017towards}
Madry, A., Makelov, A., Schmidt, L., Tsipras, D., Vladu, A.: Towards deep
  learning models resistant to adversarial attacks. arXiv preprint
  arXiv:1706.06083  (2017)

\bibitem{mathew2020monocular}
Mathew, A., Patra, A.P., Mathew, J.: Monocular depth estimators:
  Vulnerabilities and attacks. arXiv preprint arXiv:2005.14302  (2020)

\bibitem{meng2017magnet}
Meng, D., Chen, H.: Magnet: a two-pronged defense against adversarial examples.
  In: Proceedings of the 2017 ACM SIGSAC conference on computer and
  communications security. pp. 135--147 (2017)

\bibitem{nassi2020phantom}
Nassi, B., Nassi, D., Ben-Netanel, R., Mirsky, Y., Drokin, O., Elovici, Y.:
  Phantom of the adas: Phantom attacks on driver-assistance systems. IACR
  Cryptol. ePrint Arch.  \textbf{2020}, ~85 (2020)

\bibitem{papernot2016limitations}
Papernot, N., McDaniel, P., Jha, S., Fredrikson, M., Celik, Z.B., Swami, A.:
  The limitations of deep learning in adversarial settings. In: 2016 IEEE
  European symposium on security and privacy (EuroS\&P). pp. 372--387. IEEE
  (2016)

\bibitem{pei2017deepxplore}
Pei, K., Cao, Y., Yang, J., Jana, S.: Deepxplore: Automated whitebox testing of
  deep learning systems. In: proceedings of the 26th Symposium on Operating
  Systems Principles. pp. 1--18 (2017)

\bibitem{petit2015remote}
Petit, J., Stottelaar, B., Feiri, M., Kargl, F.: Remote attacks on automated
  vehicles sensors: Experiments on camera and lidar. Black Hat Europe
  \textbf{11}(2015), ~995 (2015)

\bibitem{qiu2020semanticadv}
Qiu, H., Xiao, C., Yang, L., Yan, X., Lee, H., Li, B.: Semanticadv: Generating
  adversarial examples via attribute-conditioned image editing. In: European
  Conference on Computer Vision. pp. 19--37. Springer (2020)

\bibitem{ramamonjisoa-2021-wavelet-monodepth}
Ramamonjisoa, M., Firman, M., Watson, J., Lepetit, V., Turmukhambetov, D.:
  Single image depth prediction with wavelet decomposition. In: Proceedings of
  the IEEE/CVF Conference on Computer Vision and Pattern Recognition (June
  2021)

\bibitem{ranjan2019attacking}
Ranjan, A., Janai, J., Geiger, A., Black, M.J.: Attacking optical flow. In:
  Proceedings of the IEEE/CVF International Conference on Computer Vision. pp.
  2404--2413 (2019)

\bibitem{rao2020adversarial}
Rao, S., Stutz, D., Schiele, B.: Adversarial training against
  location-optimized adversarial patches. In: European Conference on Computer
  Vision. pp. 429--448. Springer (2020)

\bibitem{sato2021dirty}
Sato, T., Shen, J., Wang, N., Jia, Y., Lin, X., Chen, Q.A.: Dirty road can
  attack: Security of deep learning based automated lane centering under
  physical-world attack. In: 30th $\{$USENIX$\}$ Security Symposium
  ($\{$USENIX$\}$ Security 21). pp. 3309--3326 (2021)

\bibitem{sato2020hold}
Sato, T., Shen, J., Wang, N., Jia, Y.J., Lin, X., Chen, Q.A.: Hold tight and
  never let go: Security of deep learning based automated lane centering under
  physical-world attack. arXiv preprint arXiv:2009.06701  (2020)

\bibitem{sato2021wip}
Sato, T., Shen, J., Wang, N., Jia, Y.J., Lin, X., Chen, Q.A.: Wip:
  Deployability improvement, stealthiness user study, and safety impact
  assessment on real vehicle for dirty road patch attack. In: Workshop on
  Automotive and Autonomous Vehicle Security (AutoSec). vol.~2021, p.~25 (2021)

\bibitem{sharif2016accessorize}
Sharif, M., Bhagavatula, S., Bauer, L., Reiter, M.K.: Accessorize to a crime:
  Real and stealthy attacks on state-of-the-art face recognition. In:
  Proceedings of the 2016 acm sigsac conference on computer and communications
  security. pp. 1528--1540 (2016)

\bibitem{shen2020drift}
Shen, J., Won, J.Y., Chen, Z., Chen, Q.A.: Drift with devil: Security of
  multi-sensor fusion based localization in high-level autonomous driving under
  $\{$GPS$\}$ spoofing. In: 29th $\{$USENIX$\}$ Security Symposium
  ($\{$USENIX$\}$ Security 20). pp. 931--948 (2020)

\bibitem{shin2017illusion}
Shin, H., Kim, D., Kwon, Y., Kim, Y.: Illusion and dazzle: Adversarial optical
  channel exploits against lidars for automotive applications. In:
  International Conference on Cryptographic Hardware and Embedded Systems. pp.
  445--467. Springer (2017)

\bibitem{song2018physical}
Song, D., Eykholt, K., Evtimov, I., Fernandes, E., Li, B., Rahmati, A., Tramer,
  F., Prakash, A., Kohno, T.: Physical adversarial examples for object
  detectors. In: 12th $\{$USENIX$\}$ Workshop on Offensive Technologies
  ($\{$WOOT$\}$ 18) (2018)

\bibitem{lm-reloc-2020}
von Stumberg, L., Wenzel, P., Yang, N., Cremers, D.: Lm-reloc:
  Levenberg-marquardt based direct visual relocalization. In: International
  Conference on 3D Vision (3DV) (2020)

\bibitem{tang2021fooling}
Tang, K., Shen, J.S., Chen, Q.A.: Fooling perception via location: A case of
  region-of-interest attacks on traffic light detection in autonomous driving.
  In: NDSS Workshop on Automotive and Autonomous Vehicle Security (AutoSec)
  (2021)

\bibitem{thys2019fooling}
Thys, S., Van~Ranst, W., Goedem{\'e}, T.: Fooling automated surveillance
  cameras: adversarial patches to attack person detection. In: Proceedings of
  the IEEE/CVF Conference on Computer Vision and Pattern Recognition Workshops.
  pp.~0--0 (2019)

\bibitem{trippel2017walnut}
Trippel, T., Weisse, O., Xu, W., Honeyman, P., Fu, K.: Walnut: Waging doubt on
  the integrity of mems accelerometers with acoustic injection attacks. In:
  2017 IEEE European symposium on security and privacy (EuroS\&P). pp. 3--18.
  IEEE (2017)

\bibitem{tsai2020robust}
Tsai, T., Yang, K., Ho, T.Y., Jin, Y.: Robust adversarial objects against deep
  learning models. In: Proceedings of the AAAI Conference on Artificial
  Intelligence. vol.~34, pp. 954--962 (2020)

\bibitem{tu2018injected}
Tu, Y., Lin, Z., Lee, I., Hei, X.: Injected and delivered: Fabricating implicit
  control over actuation systems by spoofing inertial sensors. In: 27th
  $\{$USENIX$\}$ Security Symposium ($\{$USENIX$\}$ Security 18). pp.
  1545--1562 (2018)

\bibitem{wang2019neural-NC}
Wang, B., Yao, Y., Shan, S., Li, H., Viswanath, B., Zheng, H., Zhao, B.Y.:
  Neural cleanse: Identifying and mitigating backdoor attacks in neural
  networks. In: 2019 IEEE Symposium on Security and Privacy (SP). pp. 707--723.
  IEEE (2019)

\bibitem{wang2019pseudo}
Wang, Y., Chao, W.L., Garg, D., Hariharan, B., Campbell, M., Weinberger, K.Q.:
  Pseudo-lidar from visual depth estimation: Bridging the gap in 3d object
  detection for autonomous driving. In: Proceedings of the IEEE/CVF Conference
  on Computer Vision and Pattern Recognition. pp. 8445--8453 (2019)

\bibitem{watson2021temporal-manydepth}
Watson, J., Aodha, O.M., Prisacariu, V., Brostow, G., Firman, M.: {The Temporal
  Opportunist: Self-Supervised Multi-Frame Monocular Depth}. In: Computer
  Vision and Pattern Recognition (CVPR) (2021)

\bibitem{watson-2019-depth-hints}
Watson, J., Firman, M., Brostow, G.J., Turmukhambetov, D.: Self-supervised
  monocular depth hints. In: The International Conference on Computer Vision
  (ICCV) (October 2019)

\bibitem{wimbauer2020monorec}
Wimbauer, F., Yang, N., von Stumberg, L., Zeller, N., Cremers, D.: Monorec:
  Semi-supervised dense reconstruction in dynamic environments from a single
  moving camera. In: IEEE Conference on Computer Vision and Pattern Recognition
  (CVPR) (2021)

\bibitem{wong2020targeted}
Wong, A., Cicek, S., Soatto, S.: Targeted adversarial perturbations for
  monocular depth prediction. arXiv preprint arXiv:2006.08602  (2020)

\bibitem{xiao2018generating}
Xiao, C., Li, B., Zhu, J.Y., He, W., Liu, M., Song, D.: Generating adversarial
  examples with adversarial networks. arXiv preprint arXiv:1801.02610  (2018)

\bibitem{xiao2019characterizing}
Xiao, C., Pan, X., He, W., Peng, J., Sun, M., Yi, J., Liu, M., Li, B., Song,
  D.: Characterizing attacks on deep reinforcement learning. arXiv preprint
  arXiv:1907.09470  (2019)

\bibitem{xiao2019meshadv}
Xiao, C., Yang, D., Li, B., Deng, J., Liu, M.: Meshadv: Adversarial meshes for
  visual recognition. In: Proceedings of the IEEE/CVF Conference on Computer
  Vision and Pattern Recognition. pp. 6898--6907 (2019)

\bibitem{xiao2018spatially}
Xiao, C., Zhu, J.Y., Li, B., He, W., Liu, M., Song, D.: Spatially transformed
  adversarial examples. arXiv preprint arXiv:1801.02612  (2018)

\bibitem{xu2020adversarial}
Xu, K., Zhang, G., Liu, S., Fan, Q., Sun, M., Chen, H., Chen, P.Y., Wang, Y.,
  Lin, X.: Adversarial t-shirt! evading person detectors in a physical world.
  In: European Conference on Computer Vision. pp. 665--681. Springer (2020)

\bibitem{xu2017feature}
Xu, W., Evans, D., Qi, Y.: Feature squeezing: Detecting adversarial examples in
  deep neural networks. arXiv preprint arXiv:1704.01155  (2017)

\bibitem{yamanaka2020adversarial}
Yamanaka, K., Matsumoto, R., Takahashi, K., Fujii, T.: Adversarial patch
  attacks on monocular depth estimation networks. IEEE Access  \textbf{8},
  179094--179104 (2020)

\bibitem{yan2016can}
Yan, C., Xu, W., Liu, J.: Can you trust autonomous vehicles: Contactless
  attacks against sensors of self-driving vehicle. Def Con  \textbf{24}(8),
  ~109 (2016)

\bibitem{yang2020d3vo}
Yang, N., Stumberg, L.v., Wang, R., Cremers, D.: D3vo: Deep depth, deep pose
  and deep uncertainty for monocular visual odometry. In: Proceedings of the
  IEEE/CVF Conference on Computer Vision and Pattern Recognition. pp.
  1281--1292 (2020)

\bibitem{zhang2019defending}
Zhang, Y., Liang, P.: Defending against whitebox adversarial attacks via
  randomized discretization. In: The 22nd International Conference on
  Artificial Intelligence and Statistics. pp. 684--693. PMLR (2019)

\bibitem{zhang2020adversarial}
Zhang, Z., Zhu, X., Li, Y., Chen, X., Guo, Y.: Adversarial attacks on monocular
  depth estimation. arXiv preprint arXiv:2003.10315  (2020)

\end{thebibliography}

\appendix
\renewcommand{\thesubsection}{\Alph{subsection}}
\renewcommand\thefigure{A\arabic{figure}}    
\setcounter{figure}{0}    

\renewcommand\thetable{A\arabic{table}}    
\setcounter{table}{0}    

\renewcommand\theequation{A\arabic{equation}}    
\setcounter{equation}{0}

\section*{Appendix}
\label{sec:supplemental}
 This document provides more details about our work and additional experimental settings and results. We organize the content of Appendix as follows:
 \begin{enumerate}[$\bullet$]
     \item \S\ref{sec:appx_multi}: Optimizing multiple patch regions.
     \item \S\ref{append:style_losses}: Style transfer loss terms.
     \item \S\ref{append:model_selection}: MDE model selection criteria.
     \item \S\ref{append:transfer}: Transferability evaluation.
     \item \S\ref{append:ablation}: More ablation studies.
     \item \S\ref{append:phy_setting}: Physical world experiments settings.
     \item \S\ref{append:3D_model}: 3D object detection settings.
     \item \S\ref{append:def_methods}: Defence methods and discussion.
\end{enumerate}

\subsection{Optimizing multiple patch regions}\label{sec:appx_multi}

\begin{figure}[!pb]
	\centering
	\begin{minipage}{.45\columnwidth}
		\centering
		\includegraphics[width=\columnwidth]{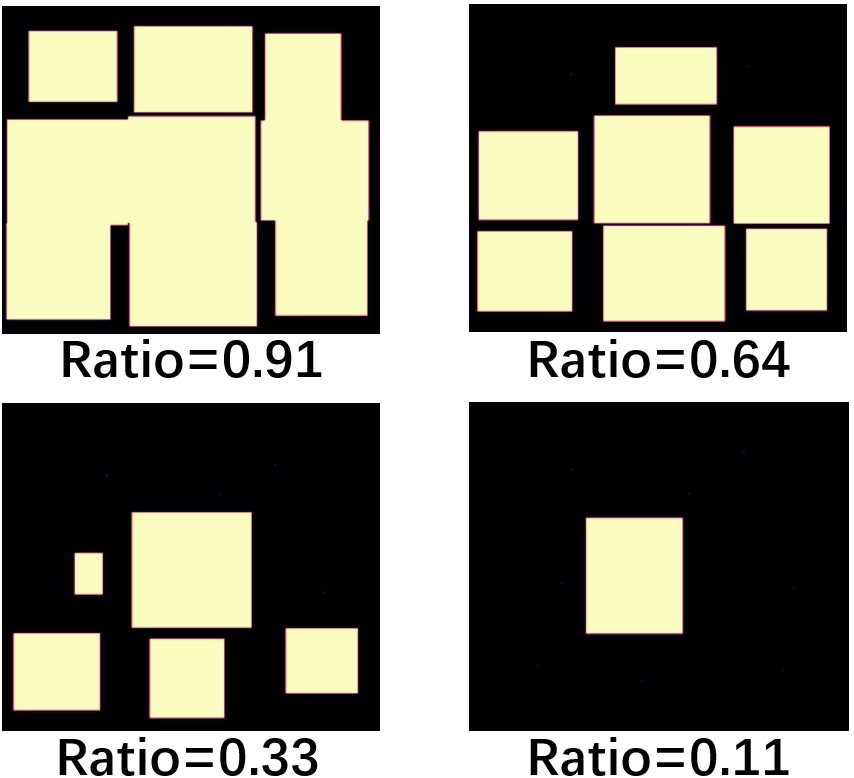}
		\caption{\small Optimization with multiple initial patches.}
		\label{fig:multi_patch_process}
	\end{minipage}%
	\hspace{0.1cm}
	\begin{minipage}{.45\columnwidth}
		\centering
		\includegraphics[width=\columnwidth]{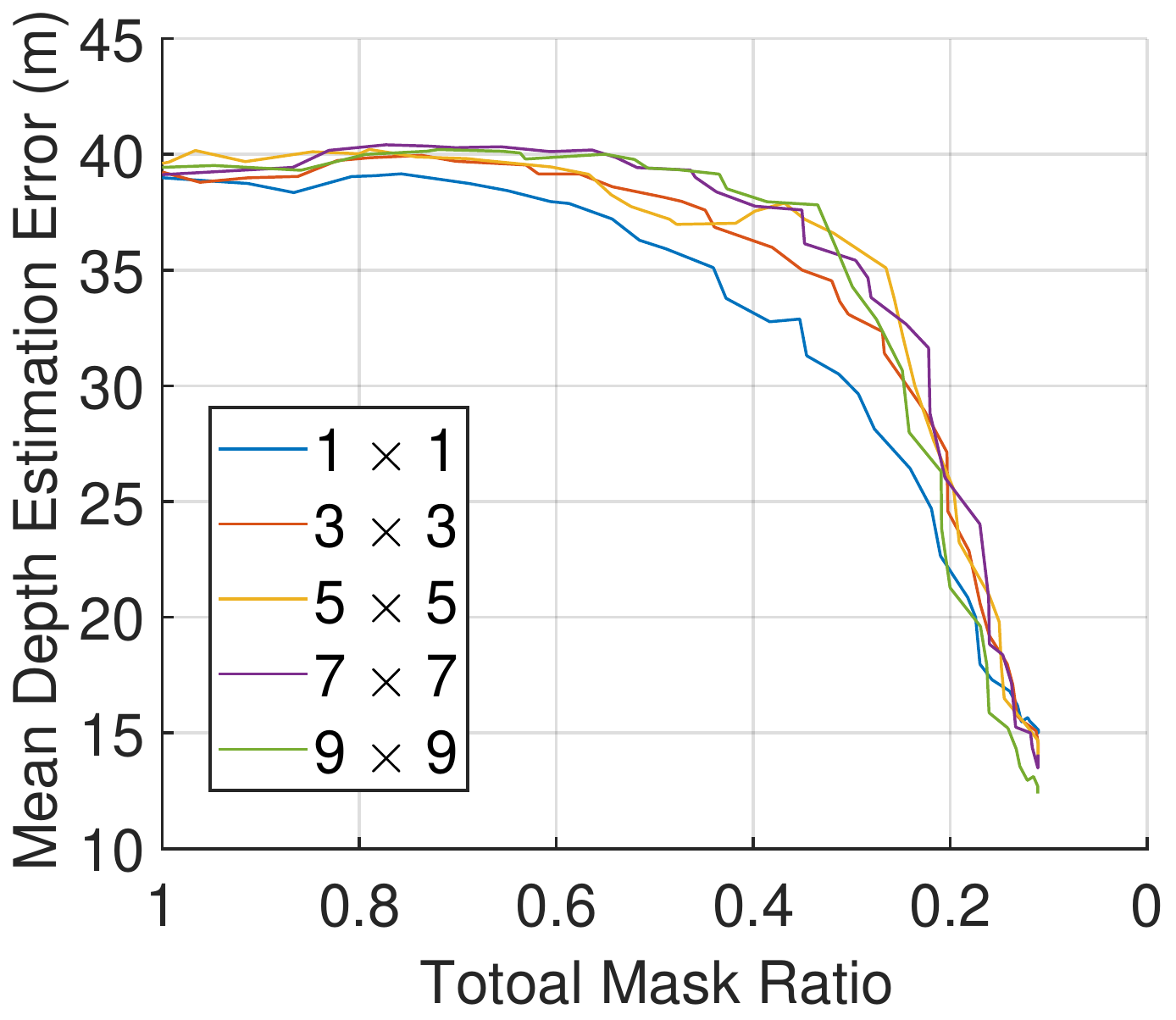}
		\caption{\small Optimization with multiple initial patches.}
		\label{fig:multi_patch_perform}
	\end{minipage}
\end{figure}

The patch region does not have to be one part. There could be multiple patches on the target object and our regional optimization technique also supports optimizing multiple patches. In this case, the final patch mask is the sum of multiple sub-masks with each sub-mask representing one rectangular region. The final mask is defined in Equation~\ref{eq:mask_region}, where $m_p^{\Theta_i}$ refers to the $i$-th sub-mask and $\Theta_i$ denotes its boundary parameters and $clamp()$ is a function to restrict the mask values in between 0 and 1. The mask loss $\mathcal{L}_m$ is also the sum of all sub-mask loss terms.

\begin{gather}\label{eq:mask_region}
	m_p^k = clamp(\sum\limits_{i=0}^k m_p^{\Theta_i}, 0, 1)
\end{gather}

Fig.~\ref{fig:multi_patch_process} gives an example of the optimization process of $3 \times 3$ initial patches. As the optimization continues, some patches are minimized and disappear, and the final patch is dominated by a single one when the target ratio is 1/9 of the vehicle's back view. We test optimizing with different initial patch setups and the result is shown in Fig.~\ref{fig:multi_patch_perform}. Each curve represents the change of attack effect as the total patch ratio decreases and $i\times j$ refers to a initial setup of $i$ rows and $j$ columns. As shown, when the target mask ratio is 1/9, different initial patch setups have similar attack performance at the end with the same total patch area. Thus, we mainly focus on optimization of a single patch ($i.e.,$ $1\times1$ setup) in our evaluation.

\subsection{Style transfer loss terms} \label{append:style_losses}

\smallskip\noindent
\noindent\textbf{Style Loss.} Let $F$ be the feature extraction network, which can be a pre-trained CNN model, $x_s$ be the style reference image, $x'$ be the adversarial patch example that we will update iteratively. The style loss is defined as the style distance between target image and adversarial example:
\begin{gather}
\small
    \mathcal{L}_s = \sum\limits_{l=1}^L\left\|G(F_l(x_s)) - G(F_l(x'))\right\|_2^2
\end{gather}
, where $F_l$ is the extracted features at the $l$-th layer of $F$ and $G$ is the Gram matrix of the deep features. $L$ is the total number of convolutional layers in $F$. 

\smallskip\noindent
\noindent\textbf{Content Loss.} The content loss is designed to preserve the content of the original image since the style loss could make the adversarial example different a lot from the original one. It is defined in Equation~\ref{eq:content_loss}:
\begin{gather}\label{eq:content_loss}
\small
    \mathcal{L}_c = \sum\limits_{l=1}^L\left\| F_l(x) - F_l(x')\right\|_2^2
\end{gather}
, where $x$ is the original image ($i.e.,$ content image). The content loss is to make sure the adversarial example and the original image have similar representation in the deep feature space. In the beginning, the adversarial example is initialized as the content image and the content loss is zero. Content loss will increase once the adversarial example is updated.

\smallskip\noindent
\noindent\textbf{Photorealism Regularization.} This term introduced in~\cite{luan2017deep} is to constrain the reconstructed image ($i.e.,$ adversarial example) to be represented by locally affine color transformations of the content image to prevent distortions~\cite{luan2017deep}, which could make the generated image more realistic. Formally, it is built upon the Matting Laplacian by Levin et al.~\cite{levin2007closed} and defined as follows:
\begin{gather}
\small
    \mathcal{L}_r = \sum\limits_{c=1}^3 V_c(x')^\top \mathcal{M}(x) V_c(x')
\end{gather}
, where $c$ represents the $c$-th color channel and $V_c(x')$ outputs the vectorized version of the $c$-th channel of the adversarial example ($i.e.,$ $V_c(x') \in \mathbf{R}^{N\times1}$, where $N$ is the number of pixels in image $x'$). $\mathcal{M}(x) \in \mathbf{R}^{N\times N}$ is a matrix only depending on content image $x$ and it represents a standard linear system that can minimize a least-square penalty function described in~\cite{levin2007closed}. We refer readers to the original article for a detailed derivation. 

\smallskip\noindent
\noindent\textbf{Smoothness loss.} This loss is designed to reduce the difference between adjacent pixels and encourage a locally smooth output image. As pointed out in \cite{sharif2016accessorize}, the smoothness term is useful in improving the robustness of a physical-world adversarial examples. The smoothness loss is defined in Equation~\ref{eq:smooth}:
\begin{gather}\label{eq:smooth}
\small
    \mathcal{L}_t = \sum\limits_{i,j}\left(x'[i,j] - x[i+1,j])^2 + (x'[i,j] - x[i,j+1])^2\right)^{\frac{1}{2}}
\end{gather}
, where $x[i,j]$ represents the pixel at $i$-th row and $j$-th column of image $x$.


\subsection{Model selection criteria} \label{append:model_selection}

MDE models can be either trained with supervised method (using ground truth depth collected by Lidar or depth camera) or unsupervised method (using video frames or stereo image pairs).  Unsupervised models are more attractive to industry because one can easily collect a large amount of training data (e.g. videos) with affordable RGB cameras or reuse existing videos at a low cost. Tesla has declared that they use a self-supervised model in monocular depth estimation\cite{tesla-self-supervised}. Hence, in our evaluation, we use three monocular depth estimation models: Monodepth2~\cite{monodepth2}, Depthhints~\cite{watson-2019-depth-hints}, and Manydepth~\cite{watson2021temporal-manydepth}. We selected these models with the following criteria.

(1)~\textit{Representativeness}. Among self-supervised monocular depth estimation models, these models are most widely used in many previous research~\cite{yang2020d3vo,ramamonjisoa-2021-wavelet-monodepth}. Monodepth2~\cite{monodepth2} is one of most successful monocular depth estimation methods. Depthhints~\cite{watson-2019-depth-hints} is an advanced model that improves performance via additional depth suggestion obtained from stereo algorithms. Manydepth~\cite{watson2021temporal-manydepth} is the state-of-the-art model that uses sequence information from multiple images to achieve better performance.

(2)~\textit{Practicality}. We focus on self-supervised monocular depth estimation models because they do not require ground truth depth data for training, which is usually collected by high-priced Lidar
sensors. In contrast, they require only monocular videos or stereo pairs collected by RGB camera(s), enabling them to collect data and train a model economically and efficiently. These techniques have already been in production-level vision-based autonomous driving systems such as Tesla Autopliot~\cite{tesla-self-supervised} and Baidu Apollo Lite~\cite{baidu_apollo_lite}. 

(3)~\textit{Open Model}. The models are publicly available. In our evaluation, we use models trained with both monocular videos and stereo pairs on KITTI dataset~\cite{Geiger2012CVPR-kittidataset} and the resolution of input images are 320$\times$1024. These models are publicly available in their project repository on GitHub~\cite{monodepth2_github,depthhints_github,manydepth_github}

\subsection{Transferability evaluation}\label{append:transfer}

\begin{wraptable}{r}{6.5cm}
\small
    \centering
    \caption{\small Transferability evaluation.}\label{tab:transfer}
    \vspace{-5pt}
    \begin{subtable}{3cm}
    \centering
    \caption{\small Across Objects}
    \label{tab:transfer_obj}
    \scalebox{0.8}{
    \begin{tabular}{cccc}
    \toprule
    \rowcolor{mygray}
     & V-A & V-B & V-C \\ \midrule
    V-A & \textbf{12.84} & 7.89 & 9.66 \\ \midrule
    V-B & 9.11 & \textbf{10.23} & 5.37 \\ \midrule
    V-C & 6.36 & 8.72 & \textbf{11.58} \\ \bottomrule
    \end{tabular}
    }
    \end{subtable}
    \begin{subtable}{3cm}
    \centering
    \caption{\small Across Networks}
    \label{tab:transfer_net}
    \scalebox{0.8}{
    \begin{tabular}{cccc}
    \toprule
    \rowcolor{mygray}
 & Many & DH & Mono \\\midrule
Many & \textbf{5.876} & 1.926 & 4.649 \\\midrule
DH & 0.524 & \textbf{10.027} & 9.037 \\\midrule
Mono & 0.304 & 1.9 & \textbf{12.84}\\\bottomrule
\end{tabular}
    }
    \end{subtable}
    
\end{wraptable}

We evaluate the transferability of our adversarial patch in two phases: the transferability across objects and the transferability across networks. For objects, we use three types of vehicles as our target objects. They are a black SUV (V-A), a blue sedan (V-B) and a grey truck (V-C). We use Monodepth2 as the target depth estimation model and generate adversarial patch for these three vehicles respectively. Then we paste each generated patch to three vehicles and evaluate the attacking performance. For the transferability across networks, we use the black SUV as target object and generate adversarial patches with three kinds of monocular depth estimation models, then we evaluate the attacking performance of each patch on the three networks respectively. Other experimental setup is the same as the effectiveness evaluation and we report the mean depth estimation error for attacks.

\begin{table}[!bp]
\vspace{-20pt}
	\centering
	\caption{\small Patches generated with different style transfer weights}\label{tab:style_transfer_lambda}
	\scalebox{0.8}{
	\begin{tabular}{ c c  c   p{1.5cm} p{1.5cm} p{1.5cm}}
		\toprule
		\rowcolor{mygray}
		$\lambda$ &Patch Image & Depth Gap &  $\mathcal{E}_d$ & $\mathcal{R}_a$  & SSIM\\ \midrule
		 0.1 & \begin{minipage}{.2\columnwidth}
			
			\centering
			\includegraphics[width=0.9\textwidth]{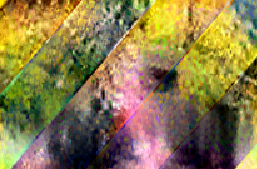}
			
		\end{minipage}
		&
		\begin{minipage}{.2\columnwidth}
			
			\centering
			\includegraphics[width=0.9\textwidth]{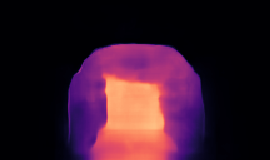}
			
		\end{minipage}
		& 11.09 & 0.373  & 0.108\\ \midrule
		 1 &\begin{minipage}{.2\columnwidth}
			
			\centering
			\includegraphics[width=0.9\textwidth]{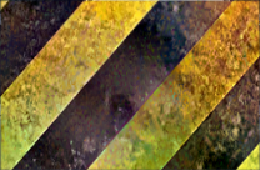}
			
		\end{minipage}
		&
		\begin{minipage}{.2\columnwidth}
			
			\centering
			\includegraphics[width=0.9\textwidth]{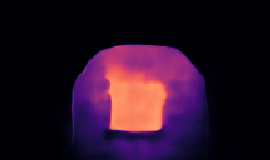}
			
		\end{minipage}
		& 7.69 & 0.246 & 0.575\\ \midrule
		10 & \begin{minipage}{.2\columnwidth}
			
			\centering
			\includegraphics[width=0.9\textwidth]{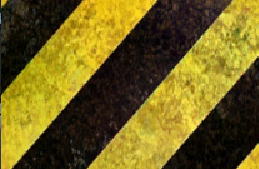}
			
		\end{minipage}
		&
		\begin{minipage}{.2\columnwidth}
			
			\centering
			\includegraphics[width=0.9\textwidth]{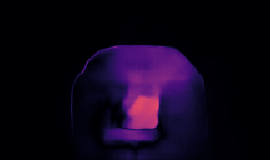}
			
		\end{minipage}
		& 1.44 & 0.003& 0.924\\ \midrule
		100 &\begin{minipage}{.2\columnwidth}
			
			\centering
			\includegraphics[width=0.9\textwidth]{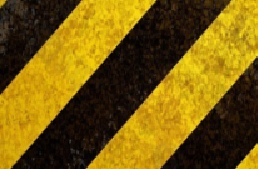}
			
		\end{minipage}
		&
		\begin{minipage}{.2\columnwidth}
			
			\centering
			\includegraphics[width=0.9\textwidth]{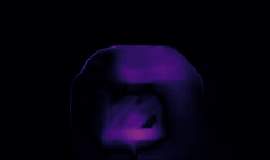}
			
		\end{minipage}
		&  0.65 & 0.001 & 0.993\\ \bottomrule
	\end{tabular}
	}
\end{table}

The result is shown in Table~\ref{tab:transfer}. In Table~\ref{tab:transfer_obj}, the first column denotes the object for which the patch is generated and the first raw denotes the object to which the patch is pasted. Observe that the adversarial patch generated from one vehicle is also effective on other vehicles, which means that our adversarial patch has a good transferability across objects. At the same time, the patch optimized for the target object has the best attacking performance compared with unmatched objects, which shows the effectiveness of our object-specific optimization. In Table~\ref{tab:transfer_net}, the first column denotes the target network used to generate the patch and the first row denotes the network used to evaluate the attack. Results on the diagonal are white-box attacks and others are black-box attacks. Observe that all three models are vulnerable to white-box attacks, which is consistent with our evaluation of attack effectiveness. For black-box attacks, Monodepth2 is the most vulnerable since patches generated from other two networks have a strong effect on it while Manydepth and Depthhints are more robust. In summary, our attack has a good transferability towards Monodepth2 but is less effective on Depth Hints and Manydepth.

\vspace{-10pt}
\subsection{More ablation studies}\label{append:ablation}

\noindent\textbf{Style transfer Weight.} We also evaluated the impact of different style transfer weight $\lambda$. Using a larger style transfer weight can generate more stealthy adversarial patterns while having worse attack performance. 
We use our default setting in all other experiments as a reference ($i.e.,$ $\lambda = 1$) and do ablation study with different style transfer weights. We use Monodepth2 and the vehicle as our target network and object. For a fair comparison, we fix the patch region to the optimized region generated in our default setting and place the vehicle at the same position on scene images in each test. Table~\ref{tab:style_transfer_lambda} presents the result. The first column is the style transfer weight parameter. The second and third columns show the generated patch image and the corresponding depth gap caused by the attack. The fourth and fifth columns are the two metrics we used to evaluate attack performance. The last column reports the Structural Similarity Index (SSIM) between the adversarial patch and the original style image, which is a metric to measure the perceptual difference between two images, ranging from 0 to 1, the higher, the more similar. As shown, using a larger style transfer weight can generate more stealthy adversarial patterns while having worse attack performance. Our default setting ($\lambda = 1$) makes a good balance between them.

\vspace{-10pt}
\subsection{Physical world experiments settings}\label{append:phy_setting}

\begin{figure}[t]
	\includegraphics[width=0.9\columnwidth]{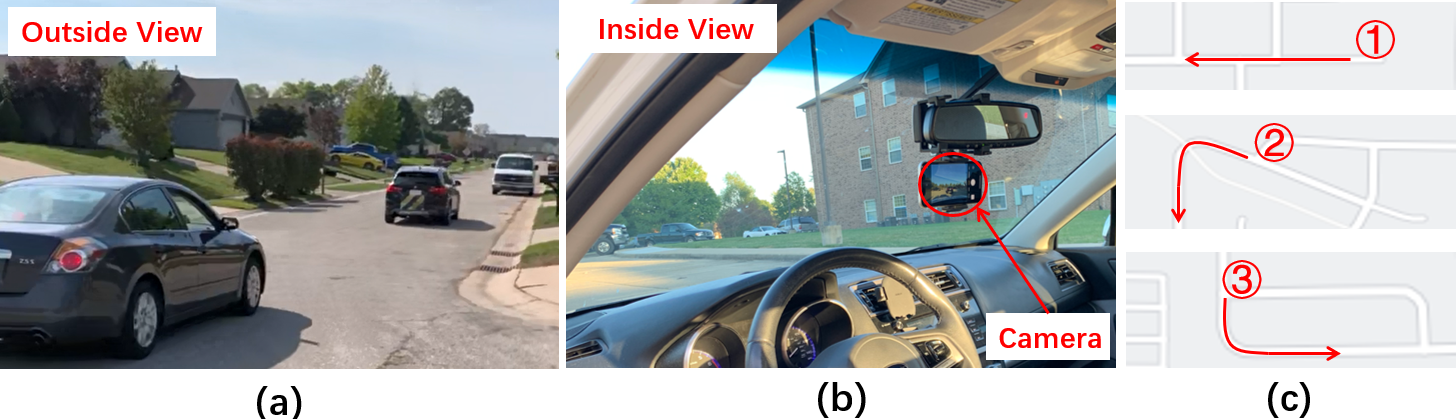}
	\centering
	\caption{\small Physical world experiments setup}
	\label{fig:physical_setup}
	\vspace{-20pt}
\end{figure}

In our physical world experiments, we use 2016 BMW X1 as our target object and Monodepth2 as the target monocular depth estimation model. We first take a photo of the vehicle's back view. Then we generate the adversarial patch with our attack method as described in \S\ref{sec:method}. Multiple background scenarios from the dataset are used in this generation process. The vehicle with the generated patch is shown in Fig.~\ref{fig:stealthy_ours}. We print the patch and paste it on the optimized region of the target vehicle to create an adversarial car. On the victim side, we use iPhone 11's back camera as the main camera of the victim vehicle. We drive the victim vehicle following the target vehicle at a distance of 7-10 meters and record the adversarial scenario while driving. Fig.~\ref{fig:physical_setup} (a) and (b) show the inside and outside view of our experimental setup. To explore the attack performance under different conditions, we drive on three routes as shown in Fig.~\ref{fig:physical_setup} (c), which involves different lighting conditions ($e.g.,$ positions of the sun and shadows), driving operations ($e.g.,$ going straight and turning), and different background scenes and objects. We drive twice on each route, with the first one a benign case and the second one an adversarial case. Specifically, in the first trip, we drive without any patch, while in the second one, we drive with the patch pasted on the target vehicle. We compare the monocular depth estimation of the benign and the adversarial case to evaluate the effect of our patch. Specifically, we report the mean depth estimation error $\mathcal{E}_{d}$ of the vehicle in both benign and adversarial cases. As we can see in Fig.~\ref{fig:Camera}, the depth ($z$) of the vehicle can be calculated with $z=fH/s$. So, given the focal length ($f$) of the camera and the  height of the vehicle in the physical world ($H$) and on the image plane ($s$), we calculate the vehicle's depth. We use this depth as vehicle's depth ground truth to calculate $\mathcal{E}_{d}$. Also, we project the depth map to pseudo-Lidar point cloud and use PointPillar network to do 3D object detection. 
\vspace{-5pt}
\subsection{3D object detection settings}\label{append:3D_model}
\vspace{-5pt}
We use PointPillars as our point cloud-based 3D detection network. The original model is trained with real Lidar data in KITTI object detection dataset. It cannot be directly applied to our pseudo-Lidar data because of their different density and distribution. Hence, we replace the real lidar data in the dataset with corresponding pseudo-Lidar data and train our own model on the new dataset. Specifically, we use Monodepth2 as the monocular depth estimation model and generate pseudo-Lidar for all images in the KITTI object detection dataset replacing original Lidar data. Then we train our PointPillar network on the generated pseudo-Lidar dataset from scratch. The training is the same as original setup and the mean average precision (mAP) of our model on the category of cars is 61.04, which is close to the performance in Apollo ($i.e.,$ 63.49). 

In each scene, we place the adversarial vehicle at a distance of 7 meters in the front of the victim vehicle, then we predict the depth of the scenario and project the depth output to 3D space generating pseudo-Lidar point cloud. Next, we use PointPillar to detect 3D objects in this scenario with the point cloud as input. 
We evaluate on the three depth estimation models and a vehicle object, and use three different target patch sizes in patch region optimization. We perform the evaluation on 100 scenarios and report the attack success rate.

\vspace{-10pt}
\subsection{Defence methods and discussion}\label{append:def_methods}

In \S\ref{sec:defence}, we evaluated five popular defencing techniques. The introduction and configuration details of these methods are as follows:

\noindent\textbf{JPEG compression}: This method uses JPEG image compressing algorithms to compress the input before feeding to the depth estimation network. The compressing operation is expected to disturb the subtle pixel-level adversarial noise and defend the adversarial attack. In our evaluation, we use Python Image Library (PIL) to apply JPEG compression to the input and select the quality level from 90 to 20. Lower quality argument means higher compression rate.

\noindent\textbf{Bit-Depth Reduction}: Typical RGB images have three channels and each channel has an 8-bit depth. Pixel value of each channel ranges from 0 to 255. Bit-depth reduction is to remap the 8-bit depth to a smaller bit depth. Lower bit-depth has smaller color space and this remap operation can also disturb the adversarial perturbation to defence the attack. In our experiments, we evaluated four smaller bit-depth cases from 5 bits to 2 bits.

\noindent\textbf{Median Blur}: Median Blur is a method to smooth the image by calculating the median of each pixel's surrounding pixels within a certain kernel size. This defence uses the smoothing effect to remove the adversarial noise. We use the median filter implemented in SciPy and use square kernel size ranging from 5 to 25 in our experiments. Larger kernel size has stronger smoothing effect.

\noindent\textbf{Gaussian Noise}: This method adds Gaussian noise on the image to disturb the adversarial perturbation since adversarial perturbation is also a kind of precise noise designed to fool the network. The Gaussian noise we add to the image is zero-mean and the standard deviation is from 0.01 to 0.1. As a reference, the image data is normalized to [0, 1]. Gaussian noise with larger standard deviation is stronger.

\noindent\textbf{Autoencoder}: The Autoencoder is a method proposed in Magnet~\cite{meng2017magnet} to defend adversarial attacks. It uses neural networks to filter out the adversarial noise which is not within the distribution of training dataset of the target model. The neural network architecture differs according to dataset and the size of input images, and in our evaluation we use architectures defined in the original work (minist and cifar10) and the architectures designed in~\cite{sato2021dirty} (Arch-1 and Arch-2) for large-size images. We train these networks with the KITTI dataset and eliminate the 100 scenes used in our evaluation to avoid in-sample evaluation.

Besides those methods we evaluated above, adversarial training~\cite{madry2017towards} is an effective method to improve the robustness of DNN models against adversarial examples.
However, traditional adversarial training is for supervised learning which requires ground truth data while the depth estimation models we target are trained in an unsupervised ($i.e.,$ self-supervised) way using videos and stereo image pairs. 
Hardening self-supervised MDE models with adversarial training effectively and efficiently is still an open problem and we leave it to future work.

Another direction is to fuse pseudo-lidar and RGB image. Since we consider fully vision-based perception system, the defense cannot include other types of sensors like Lidar, Radar or ultrasonic sensors. To avoid object detection failure, one direction is to make full use of camera frames by fusing pseudo-Lidar and RGB images. Although the point cloud of the target object is distorted by the adversarial patch, the object can still be detected in the RGB image. Fusing both sources may have more robust object detection result. Note that this cannot fundamentally defeat our attack because the object detected in the RGB image does not have depth information and the spatial relationship between the detected target object and the victim vehicle can still be wrong. Also, fusion does not solve the problem of wrong depth estimation.  

\end{document}